\documentclass[10pt,journal,compsoc]{IEEEtran}
\usepackage{stfloats}
\usepackage{tikz}
\usepackage{epsfig}
\usepackage{graphicx}
\usepackage{amsmath}
\usepackage{amssymb}
\usepackage{subcaption}
\usepackage{booktabs}
\usepackage{xspace}
\usepackage{caption}
\makeatletter
\DeclareRobustCommand\onedot{\futurelet\@let@token\@onedot}
\def\@onedot{\ifx\@let@token.\else.\null\fi\xspace}
\def\eg{\emph{e.g}\onedot} 
\def\ie{\emph{i.e}\onedot} 
\def\cf{\emph{c.f}\onedot,} 
 \def\vs{\emph{vs}\onedot}

\makeatother

\ifCLASSOPTIONcompsoc
  \usepackage[nocompress]{cite}
\else
  \usepackage{cite}
\fi
\ifCLASSINFOpdf
\else
\fi
\begin{document}
%
\title{Interpretable Deep Feature Propagation for Early Action Recognition}
%
%
%
%

\author{He~Zhao,~\IEEEmembership{Member,~IEEE,} and
        Richard~P.~Wildes,~\IEEEmembership{Member,~IEEE}
\IEEEcompsocitemizethanks{\IEEEcompsocthanksitem He Zhao is with the Department
of Electrical Engineering and Computer Science, York University, Toronto,
ON, 30332.\protect\\
E-mail: zhufl@eecs.yorku.ca
\IEEEcompsocthanksitem Richard P. Wildes is with Department
of Electrical Engineering and Computer Science, York University, Toronto,
ON.}
}

\IEEEtitleabstractindextext{%
\begin{abstract}

Early action recognition (action prediction) from limited preliminary observations plays a critical role for streaming vision systems that demand real-time inference, as video actions often possess elongated temporal spans which cause undesired latency. In this study, we address action prediction by investigating how action patterns evolve over time in a spatial feature space. There are three key components to our system. First, we work with intermediate-layer ConvNet features, which allow for abstraction from raw data, while retaining spatial layout. Second, instead of propagating features per se, we propagate their residuals across time, which allows for a compact representation that reduces redundancy. Third, we employ a Kalman filter to combat error build-up and unify across prediction start times. Extensive experimental results on multiple benchmarks show that our approach leads to competitive performance in action prediction. Notably, we investigate the learned components of our system to shed light on their otherwise opaque natures in two ways. First, we document that our learned feature propagation module works as a spatial shifting mechanism under convolution to propagate current observations into the future. Thus, it captures flow-based image motion information. Second, the learned Kalman filter adaptively updates prior estimation to aid the sequence learning process.
\end{abstract}

\begin{IEEEkeywords}
Early action recognition, action prediction, motion analysis, Kalman filtering, interpretability
\end{IEEEkeywords}
}

\maketitle

\IEEEdisplaynontitleabstractindextext

%
\IEEEpeerreviewmaketitle

\ifCLASSOPTIONcompsoc
\IEEEraisesectionheading{\section{Introduction}\label{sec:intro}}
\else
\section{Introduction}
\label{sec:introduction}
\fi
\IEEEPARstart{I}n many scenarios it is desirable to recognize actions that are being captured in video as early as feasible, rather than await their completion. Autonomous systems that need to interact with their environment in real-time can especially benefit from such action prediction (\eg autonomous vehicles and interactive robots). Nevertheless, computational research in vision-based early action recognition is limited in comparison to recognition based on processing of entire action sequences.
Action prediction shares many challenges with action recognition based on full sequences, \eg the need to deal with viewpoint and performance variations as well as  the fact that the information about the actions per se often is mixed with distracting information, \eg clutter, camera motion, occlusion and motion blur. Additional challenges present themselves for the case of prediction, \eg different action categories might share similar sub-components at different stages (\eg pushing and patting both start with stretching of arms), which makes distinctions especially difficult when only partial  information is available. More generally, incomplete executions resulting from lack of extended temporal context can lead to data that is not discriminative enough for early classification. 

\begin{figure}[t]
\small
\centering
\includegraphics[width=\linewidth]{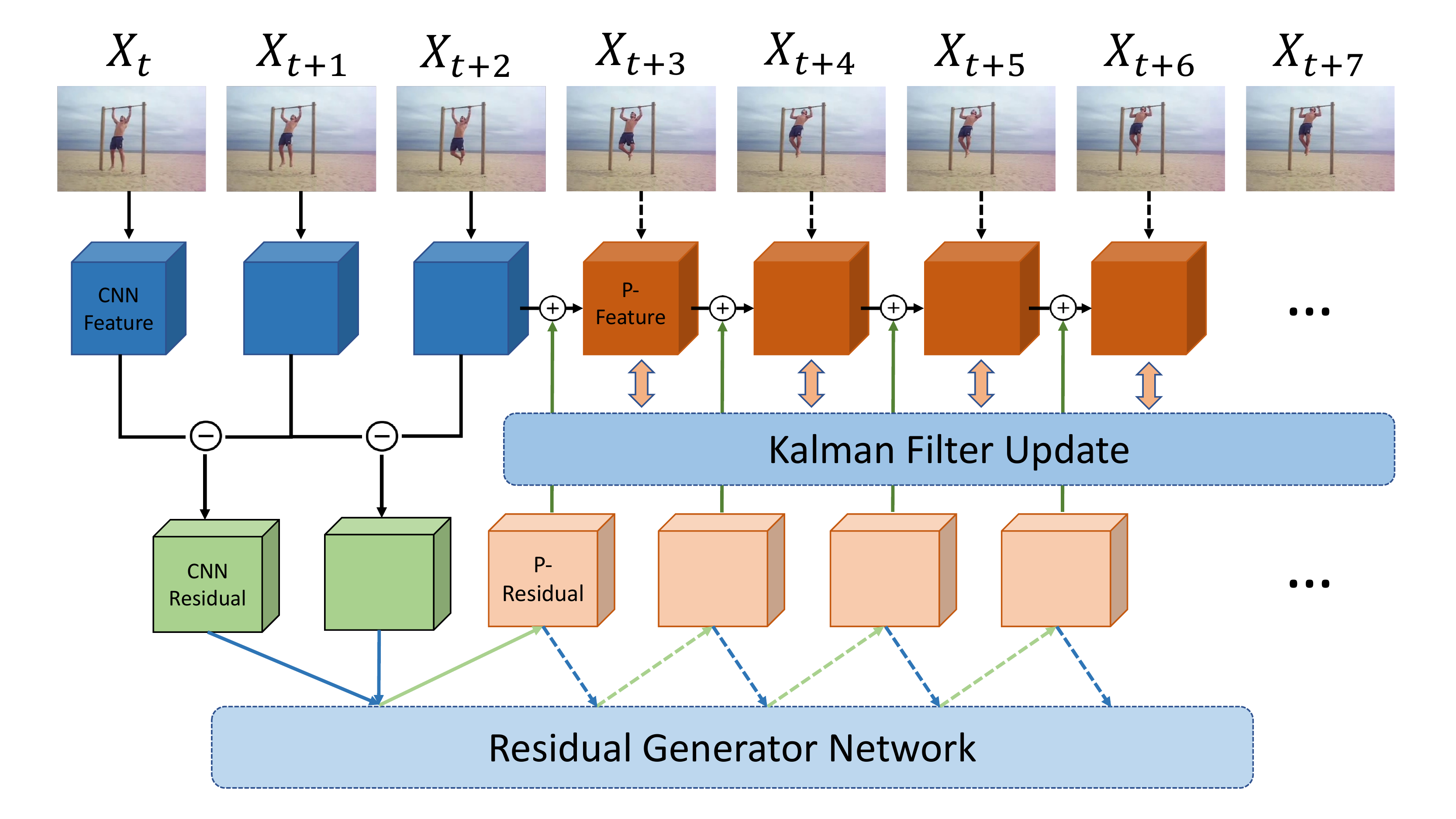}
\caption{
Overview of Proposed Feature Residual Propagation Approach to Action Prediction. Intermediate layer ConvNet features are extracted from an initial set of input frames; in the depicted example, these are given as $[X_{t}, X_{t+1}, X_{t+2}]$; subsequent frames (\eg $[X_{t+3}, \ldots, X_{t+7}]$ are not seen by the system during testing (although they are during training) and are shown here merely for context. Inital feature residuals, {\sf CNN Residuals}, are extracted via pointwise differencing of temporally adjacent feature maps. A generative model, {\sf Residual Generator Network (RGN)}, then recursively estimates future residuals, {\sf P-Residuals}. Predicted features, {\sf P-Features}, are recovered via addition of residuals to the initial reference feature map. A {\sf Kalman Filter} serves to minimize error accumulation across time. The {\sf Kalman Filter} operates across an entire video sequence during training, but only across the initially observed partial sequence during testing. Final action classification (not shown in figure) is performed with reference to both the initially observed and predicted features.
}

\vspace{-15pt}
\label{fig:teaser}
\end{figure}
Action prediction often is formulated by transferring between full video information and  partial observations. Such approaches typically overlook the rich motion patterns contained in videos, which has been demonstrated to play a pivotal role in action recognition \cite{ryoo2011human,cao2013recognize,lan2014hierarchical, kong2014discriminative}. With the recent success of deep networks on action recognition (\eg \cite{wang2016temporal, carreira2017quo, varol2018long, feichtenhofer2019slowfast}) deep representation learning based approaches offer additional possibilities. For example, one can design a temporally adaptive objective function that encourages the model to produce the correct label as early as possible \cite{sadegh2017encouraging, kong2018action}. Alternatively, one can adopt a recurrent neural network to infer recursively the next features conditioned on previous observations \cite{shi2018action, vondrick2016anticipating}. However, the fact that such approaches depend on the activation of fully-connected layers may compromise performance, as the  vectorized feature format collapses local information and contains much more noise \cite{tung2018deep}.

In response to the challenges noted above, we focus on exploring the subtle changes in spatial features across time and propose a feature Residual Generator Network (RGN) to propagate into the future. We choose intermediate level activations of a pretrained deep network for propagation (\eg final ConvLayer output, \cf \cite{zhu2017deep}), because features at such layers capture rich spatial structures \cite{zeiler2014visualizing}. 
Rather than propagate the features per se, we propagate feature residuals as they lead to a compact representation that still captures essentials of how features change over time. To ameliorate error accumulation over time, we incorporate a Kalman filter mechanism. Empirical evaluation shows that our approach yields state-of-the-art performance on three popular action prediction datasets. We also provide a detailed analysis of the representations that have been learned by our system, thereby providing interpretability. A preliminary description of this work has appeared previously \cite{zhao2019}. Most significantly, the current paper extends the previous work by providing a detailed analysis of what has been learned by all learning-based system components, something that was not considered in the previous work. Also presented is a revised Kalman filter formulation that better corrects predictions as well as new empirical results.
 Figure~\ref{fig:teaser} provides a pictorial overview of the approach. 

\vspace{1cm}
\ifCLASSOPTIONcompsoc
\IEEEraisesectionheading{\section{Related work}\label{sec:related}}
\else
\section{Related work}\label{sec:related}
\fi



\textbf{Action prediction.} Early work on video-based action prediction concentrated on use of handcrafted features. One such effort built temporally regularized discriminative models for this purpose \cite{ryoo2009spatio, ryoo2011human}. Others proposed to solve a posterior maximization on sparse feature encodings \cite{cao2013recognize} or to enforce consistency for varied temporal scales\cite{kong2014discriminative}.
More recent work has focused on deep learning. 
Some such work based prediction on action tubes over deep detectors \cite{singh2017online, singh2018predicting}.
In other work, a ConvNet with an LSTM was used to define a temporally adaptive objective function to assign labels as early as possible \cite{sadegh2017encouraging}. An alternative approach learned mappings between semantic features of full and partial videos \cite{kong2017deep, kong2018adversarial}, which was extended  with an LSTM to handle hard samples for improved performance  \cite{kong2018action}. By concentrating on relatively high-level semantic features, these approaches tend to overlook more temporally local information. To compensate for this potential shortcoming, yet other work has generated  sequential features based on current observations \cite{gammulle2019predicting, shi2018action, vondrick2016anticipating}. 
Knowledge distillation is another approach that has shown improvements on action prediction \cite{wang2019progressive}. Moreover, some recent efforts focus on discovering hard-to-discriminate details \cite {li2020hard} or exploring the relationship of multiple instances in action videos \cite{wu2021spatial}. Finally, a lone effort has extended single action prediction to group action prediction \cite{chen2020group}.

\textbf{Dynamically adaptive filters.} Convolution has became the standard operation for a wide variety of vision tasks, from low level image processing (\eg denoising) to high level semantic problems (\eg recognition). Standard convolutional approaches learn a set of fixed filters from a training set and apply them uniformly to all test data. Recent efforts have found success by conditioning the convolutional filters on input test images \cite{xue2016visual, jia2016dynamic, dai2017deformable, bako2017kernel, pouyanfar2018dynamic } for a variety of vision problems. Our work follows a similar idea to dynamically infer motion kernels from historical observations for temporal feature propagation, as action videos often exhibit diverse (sometimes even contrastive) movements that could be difficult for fixed filter banks. 

\textbf{Intermediate features.} Recent work has shown solid benefits from explicitly exploiting intermediate layer features in a multi-layer network. As examples: Intermediate features have been used for local frame aggregation \cite{lan2017deep}, building compact feature correlations \cite{diba2017deep, wang2020video}, spatial warping for real-time recognition \cite{zhu2017deep}, recovering images from various deep abstraction stages \cite{dosovitskiy2016inverting} and modulating information of distinct network branches \cite{feichtenhofer2017spatiotemporal, feichtenhofer2019slowfast}. The positive results these approaches have yielded may be explained by the fact that in comparison to fully-connected layers, intermediate layers preserve more spatial structure and thereby support finer distinctions (\eg in motion layout) as well as have fewer parameters and thereby combat overfitting. For these reasons, we build on intermediate layer features in our work on action prediction.

\textbf{Residual processing.} Residual information can play an important role in processing of redundant data even while capturing important subtle differences in an efficient fashion. MPEG-4 compression is a well established outstanding example of such processing \cite{le1991mpeg}, as is more general coarse-to-fine motion estimation (\eg \cite{Anandan89}). Recent work that exploits residual processing has considered optical-flow estimation \cite{ranjan2017optical}, image denoising \cite{jiao2017formresnet}, video artifact removal \cite{lu2018deep} and action recognition \cite{wu2018compressed}. Our approach to action prediction provides a novel use of residual processing. 

\textbf{Deep Kalman filtering.} Temporal models are useful tools for video understanding; the Kalman filter \cite{kalman1960new} is one such example,  which is well known for sequence modelling. It combines the idea of data assimilation with state-space representations and recently has been adapted with deep networks for video action detection \cite{dave2017predictive}, video prediction \cite{guen2020disentangling}, tracking \cite{haarnoja2016backprop} and temporal regularization \cite{coskun2017long}. Yet, existing work lacks sufficient understanding of why deep Kalman structure helps. Our work also uses a learnable Kalman filtering to assist the sequence training process and we provide detailed analysis on the working mechanism of the deep Kalman filter. In particular, we demonstrate that it operates as an adaptive version of the schedule sampling training technique.

\textbf{Interpretable deep learning.} A variety of techniques have been developed for understanding the operations and representations that are learned by deep networks, which otherwise remain opaque. Early work resorted to visualizing what has been learned to interpret deep networks, e.g., by showing the learned convolutional kernels \cite{zeiler2014visualizing, tran2015learning}. A parallel line of research focuses on visualizing the heat map relating the input image and output activations \cite{zhou2016learning, feichtenhofer2020deep}. 
Some others choose to fit traditional filters (\eg Gabor filters \cite{gabor1946}) to the deep framework and show that the training results in these traditional filters being learned automatically \cite{luan2018gabor,hadji2020convolutional}.  
Other recent efforts emphasize dissecting deep models by either identifying the functionality of each neuron \cite{bau2020understanding} or unrolling layers into consecutive single units \cite{monga2021algorithm}.
Our work combines visualization with analytic modelling. We visualize learned components, provide analytic interpretations and further empirically evaluate these interpretations to show the correlation between learned warping kernels and image motion information. As noted above, we also provide an explanation of our deep Kalman filter's operation. 

\vspace{1cm}
\ifCLASSOPTIONcompsoc
\IEEEraisesectionheading{\section{Technical approach}\label{sec:technical}}
\else
\section{Technical approach}\label{sec:technical}
\fi

\subsection{Overview}
We seek to predict the correct action label, $y$, given the initial portion of a partially observed video, $X_{1:k}, $ where $k$ represents the $k$th frame of a video that in total has $K$ frames. The key ingredient in support of our goal is an effective approach for  propagating the information contained in initially observed consecutive frames $X_{1:k}$ to unobserved $X_{k+1:K}$. The video action label, $y$, is then recovered via classification of the entire concatenated sequence $X_{1:K} = Cat\{ X_{1:k},  X_{k+1:K} \}$. Follow existing methods, we define the term \textit{observation ratio}, $g$, as the fraction of the observed frame set, $X_{1:k}$, to the full set, $X_{1:K}$. We present results from experiments with $g \in [0.1, 1.0]$.

Rather than predict future frames per se, we instead predict intermediate layer features of a ConvNet trained for action recognition. We are particularly interested in intermediate layer features, because features at such layers enjoy a level of abstraction from the raw data that focuses on action relevant components, even while preserving spatial layout to capture relations between action components as well as scene context. 

We decouple the prediction process into two steps: feature residual  propagation and feature reconstruction.
As discussed in Section~\ref{sec:related}, feature residual information previously has been used as a convenient proxy for full data observations as it retains salient changes to objects and motions, even while reducing redundancy entailed in explicit representation of non-changing portions of observed data. Here, we advance the application of residual extraction and processing in the domain of ConvNet features to yield a novel framework for action prediction. 

For illustrative purposes, we use the TSN architecture for initial feature extraction and final classification, because of its generally strong performance on action recognition  \cite{wang2016temporal}. While we use the TSN features and classifier, our prediction does not rely on the specifics of that approach and therefore should be more widely applicable to action prediction.




\subsection{Feature residuals}

Given a partially observed video with a set of frames $X_{1:k}$, let (mid-level) features extracted at time $t$ be denoted as $\mathbf{d}_t \in \Re^{C \times W \times H}$, with $C$ the number of feature channels, $W$ the feature map width and $H$ the feature map height. Temporal feature residuals at time $t$ are then calculated via pointwise differencing along each channel
\begin{equation} \label{eq:1}
\mathbf{r}_{t}\big\rvert_ c = \mathbf{d}_{t}\big\rvert_ c - \mathbf{d}_{t-1}\big\rvert_ c, \hspace{10pt} 2 \leq t \leq k, 1 \leq c \leq C
\end{equation}
where $\big\rvert_c$ indicates application to channel $c$, \ie the value at spatial position $(w,h)$ in channel $c$ at time $t-1$ is subtracted from the value at time $t$ and assigned to the residual, $\mathbf{r}_t \in \Re^{C \times W \times H}$, at the same spatial position and channel. 
Owing to the differencing operation, the cardinality of the set of calculated residuals, $\{ \mathbf{r}_{2:k}  \}$, is one less than the set of features, $\{ \mathbf{d}_{1:k}  \}$.

From the  limited feature set $\{ \mathbf{d}_{1:k}\}$ and their residuals set $\{\mathbf{r}_{2:k}\}$, we seek to recover the feature representation of $\{ \mathbf{d}_{k+1:K}\}$. To achieve this result, we proceed in two steps. First, we recursively generate feature residuals $\{\mathbf{r}_{k+1:K}\}$ via appeal to a feature Residual Generator Network (RGN).
Second, we sequentially add the residuals to the features that have been observed or generated so far to reconstruct features into the future according to
\begin{equation} \label{eq:2}
 \mathbf{d}_{t+1} =  \mathbf{d}_{t} + \mathbf{r}_{t+1}, \hspace{10pt} k \leq t \leq K-1. 
\end{equation} 
In Figure~\ref{fig:teaser}, {\sf P-Residuals} and {\sf P-Features} are used to distinguish predicted residuals and features, resp. In the next subsection, we define our feature residual generator.


\begin{figure}[t!]
\includegraphics[width=\linewidth]{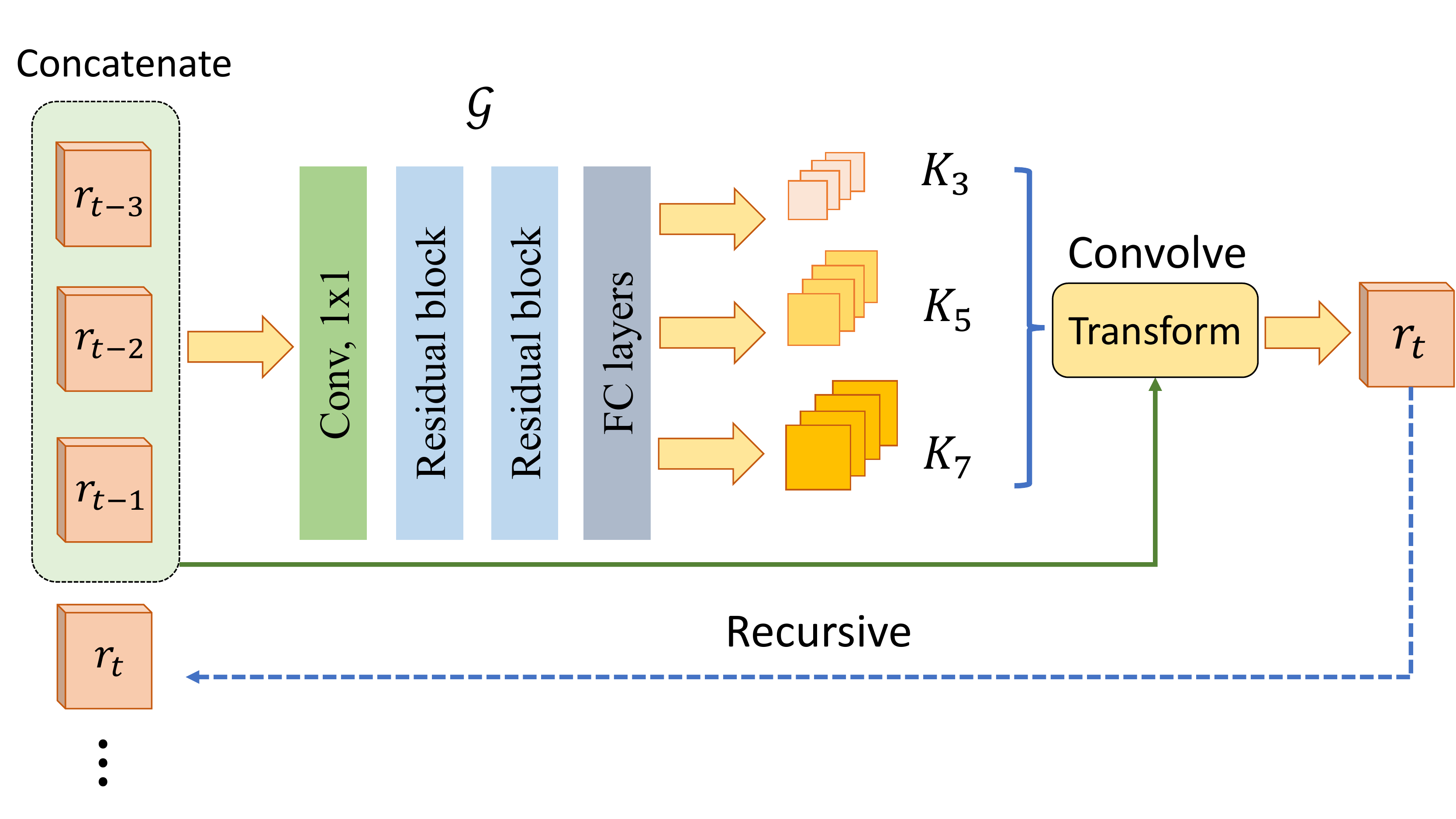}
\caption{
Temporal Extrapolation Residual Generator Network (RGN) for Predicting Next Time Step Residual. Our model recursively generates motion kernels, $K_n$, using a ConvNet, $\mathcal{G}$,  based on a short historical temporal window, $m=3$, and performs transformations in a convolutional fashion on the most recent residual. The newly generated residual joins the ongoing prediction sequence until the end of the desired sequence. Subscripts $n$ specify kernel size (\ie $n\times n$). Convolutions on the residuals are performed on a channel-by-channel basis; so, multiple kernels are depicted for each $n$.
}
\vspace{-15pt}
\label {fig:residual_motion_network}
\end{figure}

\subsection{Residual Generator Network (RGN)} 
\label{subsec:RGN}


Our Residual Generator Network (RGN) is schematized in Figure~\ref{fig:residual_motion_network}. At its core is a kernel motion transformation, $\mathcal{G}$. 
Given a set of stacked temporal observations, $\mathcal{G}$ produces a set of kernels, $\{K_n\}$, that can be convolved with the most recent residual input to predict the next (unobserved) result.  We choose the kernel motion transformation because it has proven useful in synthesis of future intensity frames \cite{finn2016unsupervised,reda2018sdc}, can be applied with various kernel sizes, $n\times n$, to capture multiple motion scales and has lower complexity than  its deep regression counterpart \cite{vondrick2016anticipating}. 


We generate motion kernels for each channel, $c$, with multiple sizes, $n \times n$,
 according to
\begin{equation} 
	\label{eq:5}
	K_{n} = \mathcal{G} (\mathbf{r}_{t}, \mathbf{r}_{t-1}, \ldots, \mathbf{r}_{t-m}\, | \, \mathbf{r}_{t-m-1}, \ldots, \mathbf{r}_{2}; \theta_{f})\big\rvert_c,
\end{equation}
where $\mathcal{G}$ is a ConvNet with learnable parameters, $\theta_f$,  that inputs residuals over its current observation window, $m$, but through its recurrent application depends on the entire history of residuals and thereby follows the Markov-Chain conditional distribution. Each of these kernels is normalized via the $l_2$ norm to avoid abnormal temporal changing, \cf \cite{finn2016unsupervised}. The architecture of $\mathcal{G}$ is depicted in Figure~\ref{fig:residual_motion_network}, with implementation details provided in Section~\ref{sec:implementation}. Subsequent to kernel generation, for each channel, $c$, we apply the kernels to the current residual $\mathbf{r}_t$ and average the results to predict the next time step residual
\begin{equation} 
	\mathbf{r}_{t+1}\big\rvert_c  =  \dfrac{1}{N} \sum_{n=1}^{N} K_{n} \circledast \mathbf{r}_{t}\big\rvert_ c, \label{eq:generator}
\end{equation}
where $\circledast$ stands for convolution. Based on preliminary experiments we use $N=3$, with $n\in \{3, 5, 7\}$.

\begin{figure}[t!]
\includegraphics[width=\linewidth]{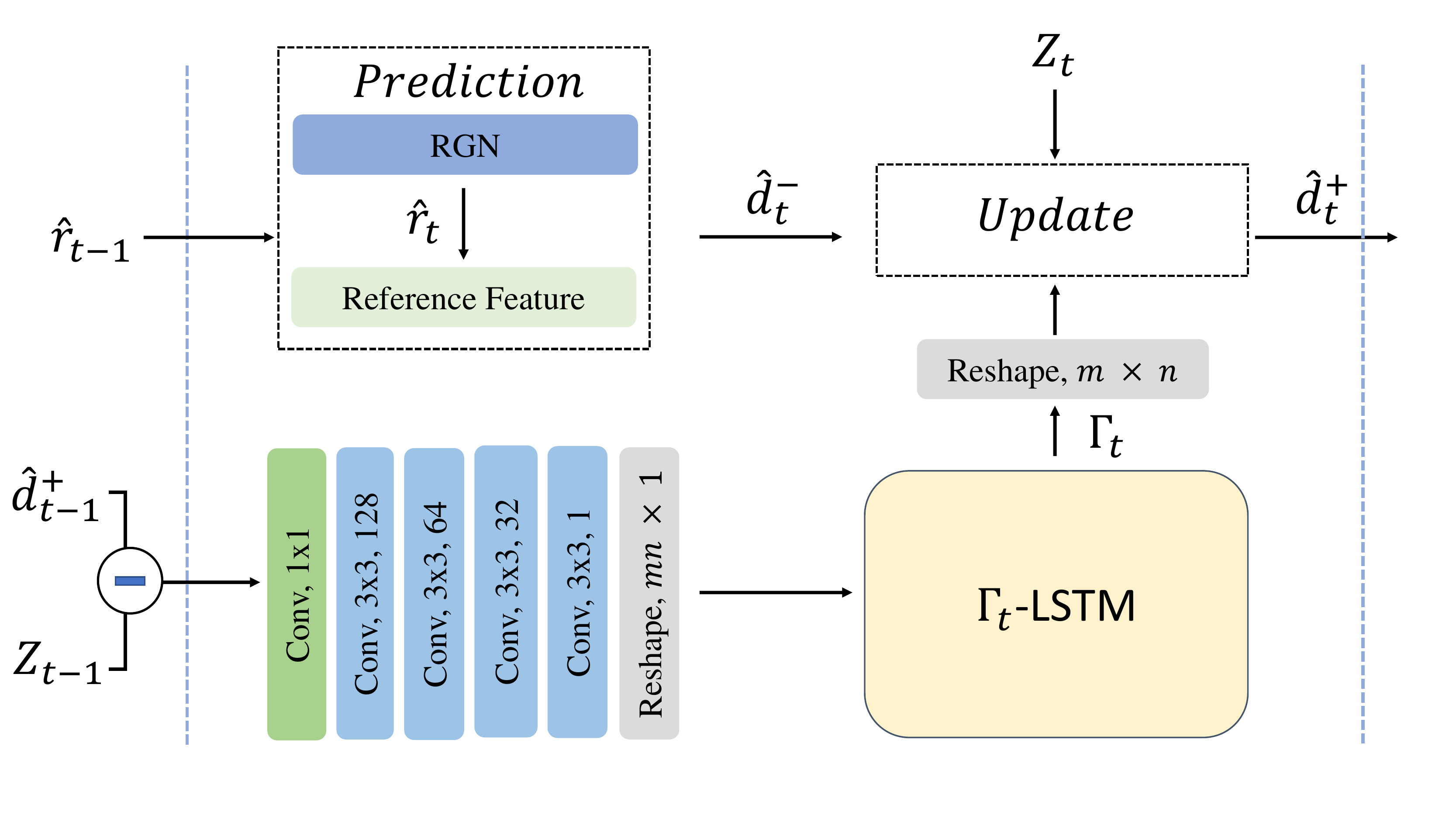}
\caption{
Depiction of Kalman update procedure. Prior estimation of feature $\hat{\mathbf{d}}_{t}^{-}$ is updated with Kalman gain $\Gamma_{t}$. The transition of $\Gamma_{t}$ is modeled by a ConvNet with  LSTM ($\Gamma_t$-{\sf LSTM}) across time. At each time step, $\Gamma_{t}$ corrects $\hat{\mathbf{d}}_{t}^{-}$ with observed measurement $Z_{t}$ and produces posterior $\hat{\mathbf{d}}_{t}^{+}$ for next time step inference. 
}
\vspace{-15pt}
\label{fig:kalmanFiltering}
\end{figure}

\subsection{Kalman filter correction} \label{subsec:KF}
Recent approaches to sequential feature generation prefer decomposing multi-step prediction into single-step prediction for training and apply the same model recursively for testing. Owing to error accumulation, such approaches often lead to quality degeneration as the sequence becomes longer. Current time-series optimization methods (\eg BackPropagation Through Time (BPTT)) lack the ability to inject mid-stage supervision during optimization; thus, errors in initial stages negatively impact the following results. To avoid such scenarios, we incorporate a Kalman filter \cite{kalman1960new} into our approach, \cf \cite{coskun2017long, lu2018deep}; see Figure~\ref{fig:kalmanFiltering}.

The Kalman filter recursively estimates an internal state from a time  series of measurements via alternating $\textit{Predict}$  and $\textit{Update}$ steps along the temporal axis. In our case, the internal state corresponds to the features recovered from the predicted residuals according to \eqref{eq:2}, while $\textit{Predict}$ is formulated as the RGN defined in Section~\ref{subsec:RGN} and $\textit{Update}$  is formulated as
\begin{equation} \label{eq:3}
\hat{\mathbf{d}}_{t}^{+} = \hat{\mathbf{d}}_{t}^{-} + \Gamma_{t} (Z_{t} - \hat{\mathbf{d}}_{t}^{-}),
\end{equation}
where\hspace{-2pt} $\hat{}$ \hspace{-2pt}distinguishes the estimated as opposed to groundtruth values, $Z_t$ is the real measurement at time $t$ (groundtruth framewise feature), $\mathbf{d}_{t}^{-}$ is the prior estimate, \eqref{eq:2}, and $\Gamma_{t}$ is the Kalman Gain. The $\textit{Update}$ corrects the current prediction by balancing the observed measurement, $Z_{t}$, and prior estimate, $\hat{\mathbf{d}_{t}^{-}}$, and the gain, $\Gamma_{t}$, defined as
\begin{equation} \label{eq:4}
\Gamma_{t} = \psi (Z_{t-1} - \hat{\mathbf{d}}_{t-1}^{+}; \theta_{z}).
\end{equation}

In the classic definition, the Kalman gain is estimated from a ratio of the process and measurement noise, both of which are pre-defined by prior knowledge of physical aspects of the system being modelled (e.g., environmental factors, like air resistance). Such noise factors can be considered as the source of prediction errors and are readily modelled under Gaussian assumptions.

However, this design is not feasible in our work, as the size of the action states is too large, \ie every feature point is viewed as an individual state variable and the employed features are large in size,  $(28, 28, 192)$. Further, it is difficult to obtain useful prior knowledge of errors when using deep networks.
Instead, we treat the Kalman gain as the output of a nonlinear function, $\psi$, of the \textit{difference} between the real and predicted features, $Z_{(t-1)} - \hat{d}_{(t-1)}^{+}$, which naturally reflects the prediction error. We realize $\psi$ as a ConvNet with an LSTM and learnable parameters, $\theta_z$,  \textit{cf}. \cite{coskun2017long}. The architecture 
 is depicted in Figure~\ref{fig:kalmanFiltering}, with details provided in Section~\ref{sec:implementation}. Note that this specification of the Kalman gain differs from that in our earlier work \cite{zhao2019}, which more simply directly input both $  Z_{(t-1)}$ and $\hat{d}_{(t-1)}^{+}$ without differencing, to the ConvNet and thereby more poorly captured the desire for the update to be conditioned on the prediction error. This new Kalman gain is named \textbf{KF-2} in the following.


We explicitly incorporate the Kalman filter \textit{Update} step into the training of the RGN, where correction happens after the estimate of $\hat{\mathbf{d}}_{t}^{-}$ is obtained, as depicted in Fig.~\ref{fig:kalmanFiltering}. The corrected feature $\hat{\mathbf{d}}_{t}^{+}$ is subsequently used for $t+1$ prediction and loss computation thereafter. During training, the Kalman filter has access to true observations, $Z_t$ throughout the video. In testing, however, the Kalman filter only has access to true observations up through the final input partial observation, $X_k$, and is only applied through that point, as detailed in Section~ \ref{subsec:progress_level}. We find that the instantaneous correction offered by the Kalman filter helps stablize long-term inference, as documented in Sec.~\ref{subsect:ablation_termpoal_model}.

\subsection{Learning scheme}
In our approach, there are two sets of trainable parameters, $\theta_{f}$ and $\theta_{z}$, that are associated with the kernel motion generator, $\mathcal{G}$, of the  residual generative network 
 and the Kalman gain transition, $\psi$, resp. 
Both sets of parameters are trained using backpropagation to minimze loss objective functions. We adopt a two stage training strategy that initially learns the $\theta_{f}$ values and subsequently learns the $\theta_z$ values, while also refining the $\theta_f$ values. 
We first train $\theta_f$ because it is more central to our overall approach in performing the essential prediction, rather than the correction. This design choice conforms to the standard Kalman filter paradigm that presupposes a sane transition module and a corrective module built on rational prior estimates \cite{kalman1960new}.  Nevertheless, ultimately the prediction and correction must work together; so, $\theta_f$ and $\theta_z$ are trained jointly  in our second stage.

The parameters $\theta_f$ are optimized with respect to four losses. The first loss pertains to the residuals 
 \begin{equation} \label{eq:loss1}
 \mathcal{L}_{2}^{res}  (\theta_{f}) = || \mathbf{r}_{t} - RGN(\mathbf{r}_{t-1}, \mathbf{r}_{t-2}, ..., \mathbf{r}_{t-m}; \theta_{f} )||^{2}_{2}
\end{equation}
where $m$ is the temporal window size. (In \eqref{eq:loss1}, note that $\mathcal{G}$ is embedded in $RGN$, but here we suppress the recursive dependence 
on all previous residuals beyond the current observation window that was given 
 in \eqref{eq:5} for the sake of compactness of notation.)
The second loss pertains to the features
\begin{equation} \label{eq: real_l2}
 \mathcal{L}_{2}^{feat}  (\theta_{f}) = || \mathbf{Z}_{t} - \hat{\mathbf{d}}_{t}^{-}||^{2}_{2} = || \mathbf{d}_{t} - (\hat{\mathbf{d}}_{t-1} + \hat{\mathbf{r}}_{t})||^{2}_{2}.
\end{equation}
As reported elsewhere \cite{mathieu2015deep, villegas2017decomposing, byeon2018contextvp}, $\mathcal{L}_{2}$ works under the Gaussian assumption that data is draw from a single parameterized Gaussian distribution and thus produces blurry outcomes. To counter this shortcoming, we include an additional two losses by applying the  \textit{Gradient Difference Loss} \cite{mathieu2015deep}, which emphasizes high frequency content, on both the features and residuals to yield
\begin{equation} \label{eq:sobel1}
 \mathcal{L}_{gdl}^{res} (\theta_{f}) = || \frac{\partial}{\partial x}\left(\mathbf{r}_{t}  - \hat{\mathbf{r}}_{t}\right) ||^{2}_{2}  
				+ ||\frac{\partial}{\partial y}\left(\mathbf{r}_{t}  - \hat{\mathbf{r}}_{t}\right) ||^{2}_{2}
\end{equation}
and
\begin{equation} \label{eq:sobel}
 \mathcal{L}_{gdl}^{feat} (\theta_{f}) = || \frac{\partial}{\partial x}\left(\mathbf{Z}_{t}  - \hat{\mathbf{d}}_{t}^{-}\right) ||^{2}_{2}  
				+ ||\frac{\partial}{\partial y}\left(\mathbf{Z}_{t}  - \hat{\mathbf{d}}_{t}^{-}\right) ||^{2}_{2}.
\end{equation}
The overall objective function for $\mathcal{G}$ is defined as
\begin{equation} \label{eq:RGNloss}
 \mathcal{L}_{2}^{\mathcal{G}}  (\theta_{f}) = \lambda_{1} \mathcal{L}_{2}^{res} +  \lambda_{2} \mathcal{L}_{2}^{feat} +  \lambda_{3} \mathcal{L}_{gdl}^{res} + \lambda_{4} \mathcal{L}_{gdl}^{feat},
\end{equation}
with the $\lambda_i$ scalar weighting factors. Note that during the first stage of training, the Kalman filter would not be operating, as it has yet to be trained.

After training the RGN parameters, $\theta_f$,  the Kalman gain parameters, $\theta_z$, are trained, while the $\theta_f$ parameters values are refined to yield a joint optimization. 
Now, there are only two losses, both pertaining to the features, $\mathbf{d}$, because that is where the Kalman filter operates. The losses are analagous to \eqref{eq: real_l2} and \eqref{eq:sobel}, except that they are calculated on the updated posterior $\hat{\mathbf{d}_{t}^{+}}$ according to
\begin{equation} \label{eq:fine-tune}
 \mathcal{L}_{2}  (\theta_f, \theta_z) =  \mathbf{\alpha} \mathcal{L}_{2}^{feat}  (\hat{\mathbf{d}_{t}^{+}}; \theta_f, \theta_z) + \mathbf{\beta} \mathcal{L}_{gdl}^{feat} (\hat{\mathbf{d}_{t}^{+}}; \theta_f, \theta_z),
\end{equation}
with $\alpha$ and $\beta$ scalar weighting factors. 


\subsection{Unified model for all observation ratios}
\label{subsec:progress_level}
Learning a separate model for each observation ratio is not applicable in the real world. To overcome this difficulty, we design a unified training and testing strategy, as follows.

\textbf{Training.} The RGN
 begins by inputting the very first batch of residuals $[\mathbf{r}_{m}, \mathbf{r}_{m-1}, \ldots, \mathbf{r}_{2}]$ and recursively produces all the rest. In other words, our model is trained for predicting the whole sequence from the same starting point, thereby entirely ignoring observation ratios. 

\textbf{Testing.} Our testing also is invariant to observation ratio by switching modes of the Kalman filter operation so that it only corrects the estimates while data observations are available according to $g$. For example, when $g=0.6$, the proposed approach still starts from the beginning observations and propagates to the end, but in two modes: While the observation ratio is not yet reached, \ie $g \in [0.1, 0.6]$, we update predictions via reference to the observed true data by using the  Kalman filter update step, \eqref{eq:3}. After entering $g \in [0.7, 1.0]$, only prediction is performed, \eqref{eq:generator}.  

This procedure resembles tracking objects under a Kalman filter: When objects are observed, the system corrects its estimated coordinates based on true observation measurements; however, while objects are occluded, the system extrapolates possible locations based on ``up-to-now" system parameter values, \ie only the prediction step is performed.

\begin{figure*}[bt!]
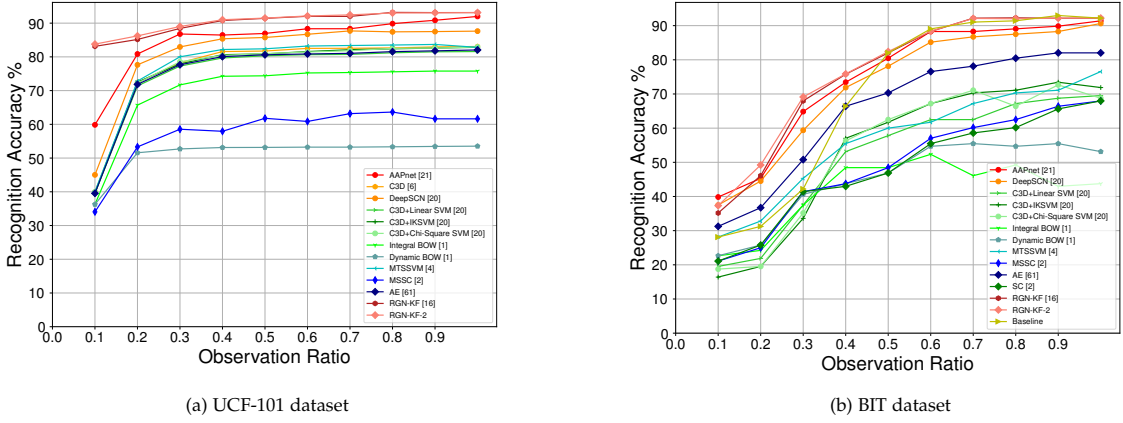

\centering
	\begin{subfigure}[b]{0.45\textwidth}
		\centering
		\scalebox{0.3}{\input{paper_components/UCF101_Accuracy.pgf}}
		\caption{UCF-101 dataset}
		\label{fig:1a}
	\end{subfigure}
		\begin{subfigure}[b]{0.45\textwidth}
		\centering
		\scalebox{0.3}{\input{paper_components/BIT_Accuracy.pgf}}
		\caption{BIT dataset}
		\label{fig:1c}
	\end{subfigure}
\caption{
Action Prediction Results on the  UCF101 and BIT Datasets at all Observation Ratios $g \in [0.1, 1]$.
}
\label{fig:result}
\end{figure*}

\vspace{1cm}
\ifCLASSOPTIONcompsoc
\IEEEraisesectionheading{\section{Empirical evaluation}\label{sec:empirical}}
\else
\section{Empirical evaluation}\label{sec:empirical}
\fi
%

\subsection{Datasets and experiment protocol}
To evaluate our approach, we choose three widely examined datasets, UCF101 \cite{soomro2012ucf101}, JHMDB21 \cite{jhuang2013towards} and BIT \cite{kong2014interactive}. UCF101 consists of 13,320 videos of 101 action categories containing a wide range of activities (\eg  sports, music and others).  JHMDB21, a subset of HMDB \cite{kuehne2011hmdb}, contains 928 videos of 21 realistic, nuanced human action categories  (\eg catching, throwing, picking). We use the provided RGB images rather than body joints of JHMDB21. 
 BIT consists of 8 classes of human interactions, with 50 videos per class. Different from the other datasets, BIT has similar behaviors of people in the initial stage of different actions (\eg they tend to be standing still) \cite{kong2018adversarial}, which leads to challenges from limited discriminatory information.

For all datasets, we use their standard train/test splits: UCF101 and JHMDB21 come with multiple train/test splits and we average over the results in our reporting, unless otherwise noted; BIT has a single train/test split, with the first 34 videos in each class for training and the rest for testing.




We present action classification accuracy as a function of observation ratio, $g$, which is the ratio of observed to total frames in a video, as used elsewhere \cite{kong2014discriminative}. Classification is always based on the concatenation of features derived from the observed frames and those that are predicted. For mid-layer features, which are the subject our propagation, we use the intermediate output of two convolutional layers and two max-poolings $\in \Re^{28\times 28 \times 192}$, unless otherwise noted. This layer is selected because empirical comparison to others generally yielded superior performance; see Section~\ref{subsect:ablation_feature_stage}. Beyond the results presented in this section, additional detailed results are provided in the supplement. 

\subsection{Implementation details}\label{sec:implementation}
To examine the propagation module with minimal influence from other factors, classifiers for chosen datasets are obtained beforehand. While a pretrained TSN model is available for UCF101 \cite{wang2016temporal}, models for JHMDB21 and BIT are not available. To adapt the TSN model to the JHMDB21 and BIT datasets, we append a simple two layer MLP classifier consisting of two hidden layers to TSN pretrained for HMDB-RGB and UCF101-Flow. For JHMDB21, the two hidden layers have 32 and 21 activations. For BIT, the two hidden layers have 64 and 8 activations. Softmax is used for final probability generation in all cases. During the training process all pretrained weights are frozen. For training of weights added for adaptation to JHMDB21 and BIT,  we randomly select 3 RGB samples or 3 Optical Flow samples (each sample has 5 frames) from the videos and get video labels by segment consensus. We employ a learning rate of 0.0001, batch size of 64, \textit{Stochastic Gradient Descent} and the \textit{Adam} optimizer. Data augmentation is the same as for the original TSN \cite{wang2016temporal}.

\textbf{Network configurations.} For the kernel generator of the RGN, $\mathcal{G}$, stacked residuals are first convolved with a $1\times 1$ kernel that reduces the feature dimension. Then, two residual convolutional blocks \cite{he2016deep} with kernel size $3\times 3$, bottleneck dimension 48 and stride 2 are used to capture temporal evolution. Subsequently, with batch and channel axis fixed, flattened spatial features are individually processed with 3 FC layers to produce $3\times 3$, $5\times 5$ and $7\times 7$ kernels. So the shape of feature map is (28, 28, 192$\times m$)-(28, 28, 192)-(28, 28, 192)-(28, 28, 192)-(9, 192), (25, 192) and (49, 192), with $m=3$ the empirically selected temporal window size. Convolution is performed on each channel. 

For Kalman Gain, $\Gamma_{t}$, a set of convolutional layers with kernel size 3x3 and stride 2 are used to capture the covariance. Each layer is appended with a ReLU layer \cite{glorot2011deep}. The shape of feature map is (28, 28, 128)-(28, 28, 64)-(28, 28, 32)-(28, 28, 1). Subsequently, the flattened feature is taken as input by $\Gamma_{t}$-LSTM to produce Kalman gain, $\Gamma_{t} \in \Re^{wh}$, which then is reshaped to $ \Gamma_{t} \in \Re^{w \times h}$, corresponding to feature map spatial dimensions. The hidden state of the LSTM has the same size as the input feature (784).
The gain is then applied according to the update, \eqref{eq:3}.

\textbf{Training strategy.} We train our model with 4 NIVDIA TITAN X GPUs, under Pytorch \cite{paszke2017automatic}. 
Training of the spatiotemporal feature residual generative network (RGN) employs the \textit{Adam} optimizer and a learning rate 0.005 with $\beta_1 = 0.9$ and $\beta_2 = 0.99$ to minimize the loss, \eqref{eq:RGNloss}.
Empirically, we set $\lambda_1, \lambda_2, \lambda_3, \lambda_4$ with ratios of 1:1:5:5, which places more emphasis on the spatial gradient rather than raw mean square values. The batch size is set to 56. Following initial training of the RGN, we fine-tune it together with the Kalman gain transition ConvNet with LSTM, $\psi$,
to minimze the loss \eqref{eq:fine-tune}. Mini-batch-Stochastic Gradient Descent is used with a learning rate of $2e^{-4}$ and exponential decay of $1e^{-5}$. $\alpha$ and $\beta$ are set empirically with a ratio of 1:5. 

For training on UCF101, we sample 30 frames from each video and use the first 3 to initialize our entire prediction system. For BIT and JHMDB21, we sample 25 frames from each video and use the first 3 to initialize our system. The TSN architecture \cite{wang2016temporal} serves to provide feature extraction and classification. We apply our system to the RGB stream for JHMDB21,  flow stream for BIT and both streams for UCF101. We make these choices following typical state-of-the-art practice on JHMDB21 (\eg  RGB features previously yielded top performance \cite{sadegh2017encouraging,shi2018action}),  BIT (\eg flow features greatly outperform spatial features \cite{kong2014interactive,kong2017deep}) and UCF101 (\eg   two-stream previously yielded top performance \cite{kong2018adversarial}).

Once features are generated, no additional modifications to TSN are needed to yield action labels. Generated features are inserted into the selected TSN mid-layer and processed up through the network tower until the MLPs produce  probability scores. Video level labeling is gathered by averaging scores from each frame. 

\subsection{Overall prediction results}
\textbf{UCF101.} Figure~\ref{fig:result} $(a)$ shows comparative results for our algorihm \textbf{RGN-KF-2} \vs various alternatives on UCF101 as well as our earlier version, \textbf{RGN-KF} \cite{zhao2019}. 
It is seen that both our approaches, \textbf{RGN-KF-2} and \textbf{RGN-KF}, outperform all others at all observation ratios, improving accuracy by $\approx$ \hspace{-3pt}3-4\% on average. The performance improvement is especially striking at lower observation ratios, \eg $g = 0.1$, where we outperform the second best (AAPnet) by 83.78\% \vs 59.85\%.
When being compared with our earlier work, \textbf{RGN-KF}, our new method uniformly surpasses it on low observation ratios, \eg $g \in (0.1, 0.2, 0.3)$, with the largest improvement at $g = 0.2$ (86.22\% \vs 85.16\%). 
Beyond that, \textbf{RGN-KF-2} performs equally well. Since the task is \textit{early} action recognition, improvements at the lowest observation ratios are of particular interest.

Notably, AAPnet also builds on TSN; however, it apparently does so less effectively than our approach does. There are likely two reasons for this state of affairs. First, AAPnet is not trained exclusively for inferring action labels, but also for adversarial learning on feature transformation, which might lessen its optimization for action prediction. Second, AAPnet more radically modifies the TSN architecture in aggregating across all frames at a given state of progress, which underlines the fact that our approach may be more adaptable to various architectures as it has less impact on their native operations.

\textbf{BIT.} Figure~\ref{fig:result} $(b)$ shows comparative results for our algorithm \vs various alternatives on BIT. It is seen that our results are equal to or better than all others, except at the lowest observation ratio, $g=0.1$. For example, compared with AAPnet, our approach achieves 69.12\% accuracy at $g=0.3$, which is 4\% higher. Notably, our \textbf{RGN-KF-2} obtains non-trivial improvements over \textbf{RGN-KF} on multiple lower ratios, \eg $g \in (0.1, 0.2, 0.3)$. Particularly at $g=0.2$, our new approach boosts the accuracy from 46.09\% to 49.96\%,
which demonstrates the solid benefit of explicitly modelling the Kalman gain from error signals, as given by, (\ref{eq:4}). Again, improvements at lowest observation ratios are of particular interest for early action recognition.

In interpreting the results on BIT it is important to recall that the beginning and ending portions of the videos tend be very similar in appearance (\eg two people standing facing one another), so that the most discriminatory information largely is limited to the intermediate portions. Correspondingly, there is a tendency for rapid performance rises after the initial portion, which levels out in the final portion. In our case, a peak performance of  92.28\% at $g=0.7$ increases that at the previous ratio by 4\%, whereas AAPnet achieves no significant increase (0.78\%) at the same stage. 

Given that we train a modified TSN architecture in adapting TSN to BIT (Section~\ref{sec:implementation}), we compare how well that modified architecture works when forced to classify on just the initially provided frames without propagation into the future. These results are shown as {\sf Baseline} in Fig.~\ref{fig:result} $b$. It is seen that by propagating into the future our approach exceeds the baseline by  large margins when $g \in [0.1, 0.4]$. For higher observation ratios, as the discriminating parts of the input videos become available to the baseline (as well as our full approach), performance becomes on par.



\textbf{JHMDB21.} The standard reporting protocol on JHMDB21 is to report recognition accuracy only when the initial 20\% of the videos are observed, \ie $g=0.2$, which we show in Table~\ref{table:jhmdb}. It is seen that our algorithm once again is the top performer, \eg \textbf{RGN-KF-2} exceeding the RBF-RNN by 6\% and \textbf{RGN-KF} by 1\%.  We also provide a baseline comparison, where we compare to classification made purely on the basis of adapting the TSN architecture to the JHMDB21 dataset, analogous to the baseline comparison provided on BIT. Once again, it is seen that our full propagation approach adds considerably to the performance of the baseline alone.
%


\begin{table}[]
\small
\centering
\begin{tabular}{l c}
\toprule
Method            & \multicolumn{1}{l}{Accuracy (\%)} \\ 
\midrule
ELSTM \cite{sadegh2017encouraging}             & 55                            \\ 
Within-class Loss \cite{ma2016learning} & 33                            \\ 
DP-SVM \cite{soomro2019online}           & 5                             \\ 
S-SVM \cite{soomro2019online}            & 5                             \\ 
Where/What \cite{soomro2016predicting}        & 10                            \\ 
Context-fusion \cite{jain2016recurrent}    & 28                            \\ 
RBF-RNN \cite{shi2018action}           & 73                            \\ 
RGN-KF \cite{zhao2019}              & 78               \\ 
RGN-KF-2              & \textbf{79}                 \\ 
Baseline              & 74                 \\ 
\bottomrule
\end{tabular}
\caption{Action Prediction Results on JHMDB21. Following the standard protocol, accuracy results are shown only for the case where initial observations are limited to the first 20\% of frames, \ie $g=0.2$.
}
\label{table:jhmdb}
\end{table}

\begin{table}[]
\centering
\begin{tabular}{l c }
\toprule
Temporal Propagation Approach & Accuracy (\%) \\ 
\midrule
\small{ConvLSTM(3x3)-128-192-\textbf{Org}}               & \small{71.1}        \\ 
\small{ConvLSTM(3x3)-128-192-\textbf{Org}-\textbf{KF}}               &  \small{73.4}        \\ 
\small{ConvLSTM(3x3)-128-192-\textbf{Org}-\textbf{KF-2}}               &  \small{74.3}        \\ 
\small{ConvLSTM(3x3)-128-192-\textbf{Res}}               &  \small{76.8}        \\ 
\small{ConvLSTM(3x3)-128-192-\textbf{Res}-\textbf{KF}}               &  \small{77.1}        \\  
\small{ConvLSTM(3x3)-128-192-\textbf{Res}-\textbf{KF-2}}               &  \small{77.3}        \\  
\small{RGN-\textbf{Org}}                    & \small{70.9}        \\ 
\small{RGN-\textbf{Org}-\textbf{KF}}                    & \small{74.4}        \\
\small{RGN-\textbf{Org}-\textbf{KF-2}}                    & \small{74.7}        \\  
\small{RGN-\textbf{Res}}                    & \small{77.4}        \\ 
\small{RGN-\textbf{Res}-\textbf{KF}}                    & \small{78.3} \\
\small{RGN-\textbf{Res}-\textbf{KF-2}}                    & \small{\textbf{79.0}}               \\ 
\bottomrule
\end{tabular}
\caption{
Accuracy results for different temporal propagation approaches on JHMDB21 split 1. \textbf{Org} denotes applying motion kernel transformation on original features, \textbf{Res} denotes residual propagation and \textbf{KF*} denotes inclusion of the Kalman filter. For ConvLSTM, (3x3), 128 \& 192 represent kernel, hidden state \& feature dimension, resp.}
\label{tab:temporal_model}
\end{table}
\begin{table*}[h]
\centering
\begin{tabular}{l c c c c c c c c c}
\toprule
          & \multicolumn{4}{c}{UCF-101} & \multicolumn{4}{c}{BIT}      & JHMDB-21 \\ 
\cmidrule(l){2-10}
\textit{observation ratio}         & 0.1   & 0.2  & 0.3   & 0.8  & 0.1   & 0.2   & 0.3   & 0.8   & 0.2      \\
\midrule
56x56x64  &    83.16   &   85.03    &   87.92    &   92.10   & 33.41 & 46.65 & 67.51 & 84.38 & 75.83    \\ 
28x28x192 &    83.78   &   86.22    &   89.01    &   93.07   & 37.33 & 49.96 & 69.12 &  88.82 & 79.00    \\ 
14x14x576 &    83.29   &   85.46   &    88.19   &   93.06  & 36.62 & 48.70 & 68.19 & 87.50 & 77.63    \\ 
7x7x1024  &    80.51   &    84.94   &    88.17   &    92.77  &  36.12&  49.22 &  67.61 & 87.50 & 77.42    \\ 
\bottomrule
\end{tabular}
\caption{
Prediction Accuracy (\%) at Various Intermediate Feature Stages with \textbf{RGN-KF-2}, Ordered by Decreasing Spatial Receptive Field Size. Observation ratio $g \in \{ 0.1, 0.2, 0.3, 0.8 \}$ for UCF101 and BIT datasets. Set $g = 0.2$ for JHMDB21.
}
\label{tab:feastg}
\end{table*}

\subsection{Influence of temporal model}
\label{subsect:ablation_termpoal_model}
In this section, we examine the influence of different temporal modeling approaches to feature propagation using JHMDB21, with ConvLSTM as an extra baseline, \textit{cf.} \cite{xingjian2015convolutional, byeon2018contextvp}; see Table~\ref{tab:temporal_model}. For both scenarios, we find that propagation on residuals is superior to propagation on raw features and the Kalman filter provides further benefits. Performance of ConvLSTM is on par with our RGN approach applied to the original features without the Kalman filter; however, for all other configurations our RGN approach performs better. Especially, our new version of the Kalman filter, \textbf{KF-2}, seems to benefit all used temporal models. Overall, we find that our full approach to temporal modeling (mid-layer convolutional feature residuals, RGN propagation and Kalman filtering) yields best performance.

\subsection{Influence of feature layers}
\label{subsect:ablation_feature_stage}
We now examine the influence of different intermediate feature spaces on prediction. We consider layers that yield feature maps of [56, 56, 64], [28, 28, 192], [14, 14, 512] and [7, 7, 1024], where [w, h, c] indicate the width, height and number of channels, resp. Table~\ref{tab:feastg} shows the results. For JHMDB21 and BIT, the [28, 28, 192] feature stage almost always achieves best results. Moreover, deeper layers,  [14, 14, 512] and [7, 7, 1024],  are more useful than the shallower layer [56, 56, 64]. This pattern of results may be explained by the earliest layer not providing adequate abstraction from the raw input, while the highest layers have lost too much distinguishing detail. 
Interestingly, for UCF101 different feature stages have less impact on accuracy. This may be due to the fact that UCF101 is generally less difficult than the other datasets, as indicated by the fact that for any given observation ratio, $g$, in Table~\ref{tab:feastg} the results on UCF101 are always better than for the others; correspondingly, the specifics of feature selection are less important. More generally, however, the results of Table~\ref{tab:feastg} support our use of intermediate layer features, especially as the prediction task becomes more difficult.

\begin{figure}[bt]
\center
\includegraphics[width=.9\linewidth]{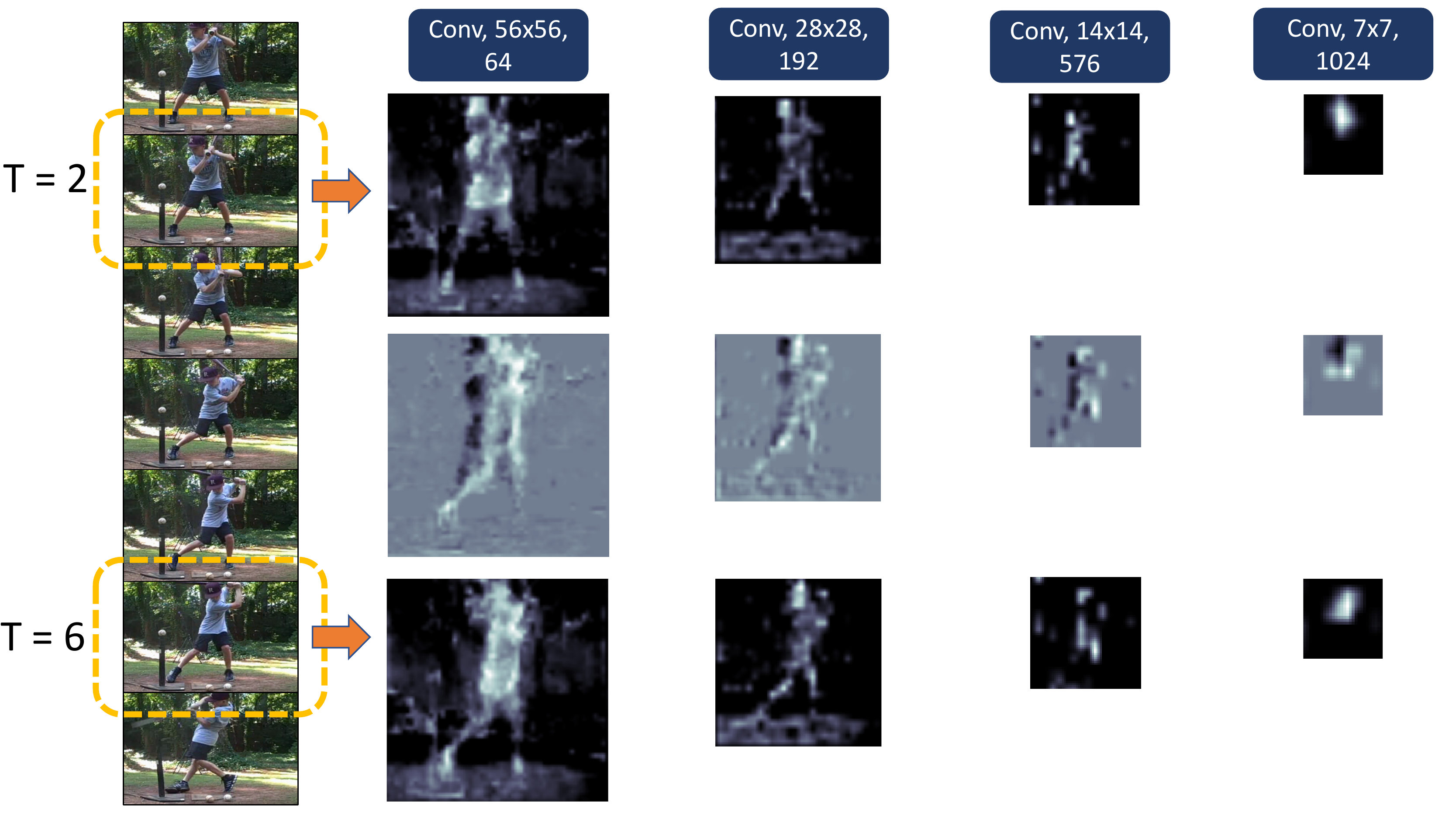}
\caption{
Visualization of Features and Residuals. The example shows a sequence of frames for the action {\em baseball-swing} along the left side. Feature maps extracted at various layers are shown along the upper/bottom row, while their residuals are shown along the middle row.}
\label{fig:fea_vis}
\vspace{-1em}
\end{figure}

\begin{figure}[h]
\center
\includegraphics[width=.9\linewidth]{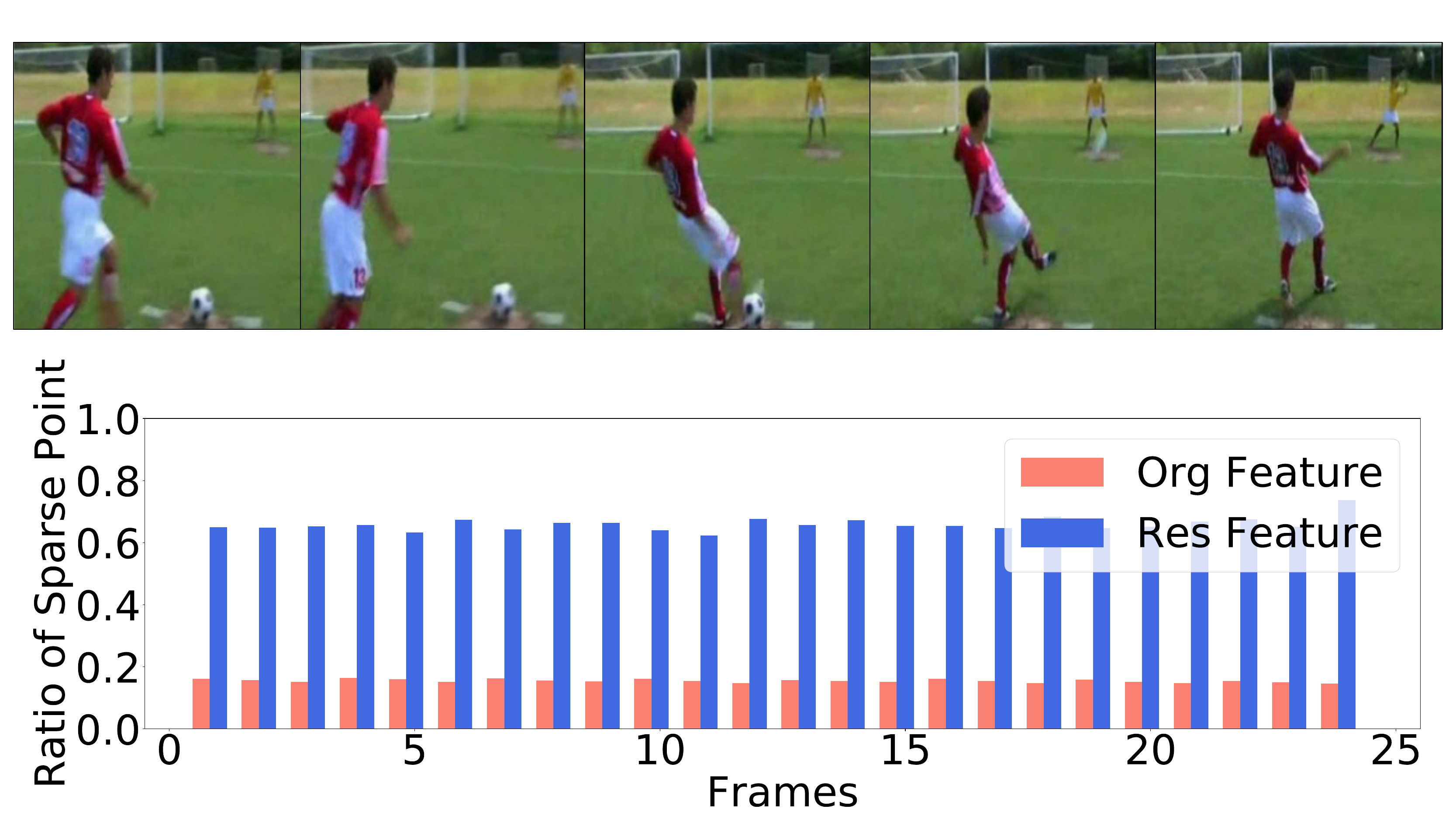}
\caption{
Sparsity Comparison of Features and Residuals. The top row shows frames from a video of a \textit{kicking} action. The bottom row shows sparsity as the ratio of near-zero value points (absolute value $< 0.01$) to total points over time. On average, residual and original feaure points are 65\% and 14\% sparse, resp.
}
\label{fig:sparse_check}
\vspace{-2em}
\end{figure}

\subsection{Visualization of feature residuals}\label{sec:visualization}
To understand further why intermediate layer features and their residuals are especially useful for action prediction, we show comparative visualizations as well as associated statistics. Figure~\ref{fig:fea_vis} provides an example from the action {\em baseball-swing}. It is seen that the earliest layer features concentrate on low-level features (\eg lines and edges) that may be too closely tied to a specific example, rather than the action class. In contrast, the latest layer features tends to lose too much distinguishing detail (\eg merely a blob in the vicinity of the actor at the top-layer). Comparatively, the mid-layer features tend to concentrate on the actor, but also delineate details of the actors parts. In comparing the raw features to their residuals, it is seen that the residuals concentrate more on temporal changes, which are good for propagating variations into the future without redundant information. Thus, intermediate layer residuals appear to capture information that is especially useful for action prediction.

The provided visualization, Figure~\ref{fig:fea_vis}, suggests that the residuals provide a more sparse (and hence compact) representation compared to the features per se. To quantify this observation, we define feature sparsity as the percentage of near-zero points (absolute value $< 0.01$) vs. total points.  Figure~\ref{fig:sparse_check} shows comparative results for original features and their residuals. It is seen that the residuals have approximately five times the sparsity of the originals, which quantitatively confirms the relative sparsity of the residuals.
Overall, both the visualizations and the quantitative analysis confirm that mid-layer feature residuals are especially information laden for action prediction.


\vspace{1cm}
\ifCLASSOPTIONcompsoc
\IEEEraisesectionheading{\section{What has been learned?}\label{sec:what_learn}}
\else
\section{What has been learned?}\label{sec:learn}
\fi

\begin{figure*}[t]
\centering
\includegraphics[width=.8\linewidth]{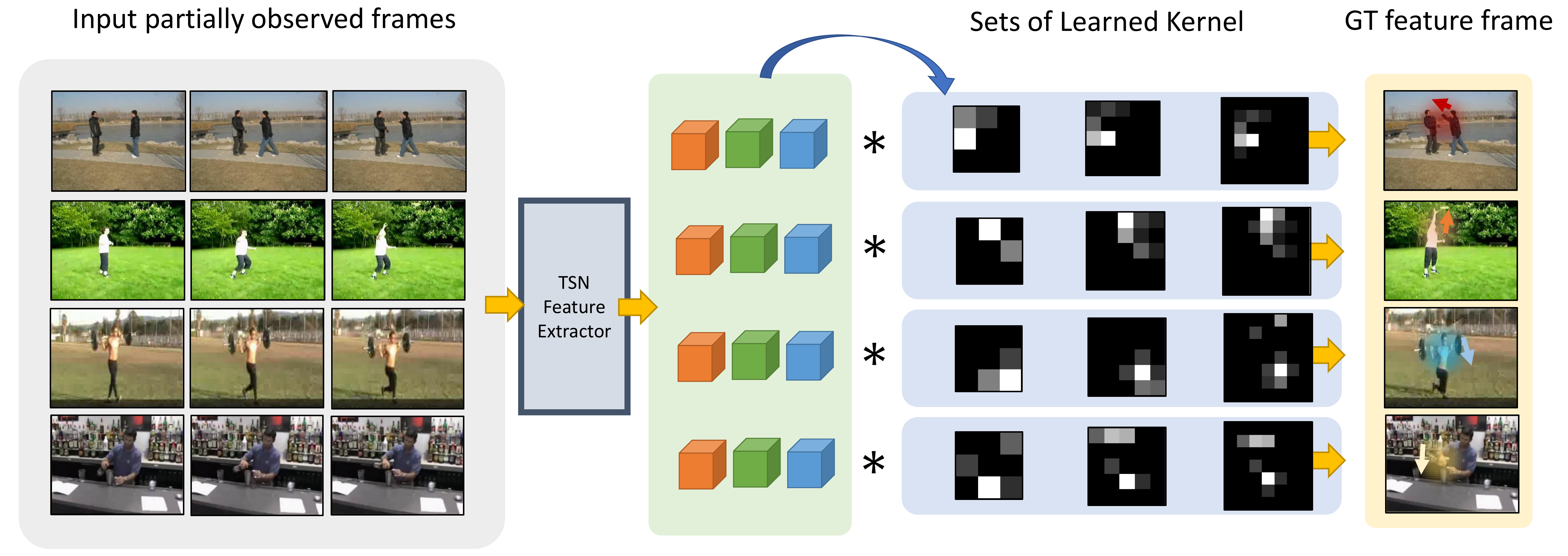}
\caption{
Visualization of Learned Motion Kernels. Left-to-right: Input frames and symbolized extracted features; learned kernels ($3\times 3$, $5\times 5$, $7\times 7$); groundtruth prediction target frames that follow the inputs, with an overlaid vector showing the dominant motion direction. Top-to-bottom are \textit{boxing}, \textit{catching}, \textit{lunges} and \textit{opening bottle} actions. Plots of the learned kernels resemble unit impulses that capture the depicted motion. 
}
\label{fig:kernel_vis}
\end{figure*}

In this section, we study the learned components of our approach in detail. Specifically, there are two major learned components: 1) dynamically inferred motion kernels, $K_{n}$, that propagate feature residuals into the future; 2) the Kalman Gain,  $\Gamma_{t}$, that updates sub-optimal prior estimates. In the following, we provide detailed examination of each using the TSN deep network as feature extractor, as it is used in our empirical evaluation of Section~4. Analysis of the features learned by TSN is available elsewhere \cite{wang2016temporal}.

\subsection{Learned motion kernels}
\subsubsection{Convolutional spatial shifting}
To understand what has been learned by the motion kernels, we being by visualizing them, as typically seen in deep network interpretation, \textit{cf.} \cite{zeiler2014visualizing,tran2015learning, feichtenhofer2020deep}; representative results are shown in Figure ~\ref{fig:kernel_vis}. In all the cases, it is seen that the learned kernel has a high intensity value displaced from its center in the direction and magnitude of the depicted motion. For example, in the top row the motion is mainly from the actor stretching his arm towards the left by approximately one pixel, with additional minor motion towards the top (denoted as red arrow). Correspondingly, the motion kernels have peak value toward the left at approximately one pixel and also some smaller values in the top and top-left. The other rows also show similar patterns of a displaced bright spot capturing the motion direction Moreover, not only is the motion direction captured, but also its magnitude: For the largest motion (second row) the displacement in the largest kernel is displaced by 3 pixels from the center, while the smaller kernels displace to the extent they can; for the intermediate magnitude motion the displacement never goes beyond 2 pixels; for the smallest displacements (top and bottom rows) the displacement is one pixel for all kernels. Interestingly, learned kernels across all actions in all datasets tend to show similar patterns. Here, it is important to note that the motion kernels are computed on a per feature channel basis and that different feature channels capture different components of the images: Some channels are better matched to foreground actors or even actor parts, while others are better matched to backgrounds. These different channels may therefore yield different motions and our per channel learning automatically follows these various motions.
 
What is the significance of such learned kernels? An explanation can had by recalling that signal shifting via convolution is captured by operating on the signal with a unit impulse that is shifted by the desired amount, \ie
\begin{equation}
g(t) \circledast \delta({t - \Delta t}) = \delta({t - \Delta t}) \circledast g(t)   = g(t - \Delta t)
\label{eq:shift}
\end{equation}
\smallskip
where $\delta(\cdot)$ denotes the unit impulse and $g(\cdot)$ denotes an arbitrary function \cite{bracewell1986}. In this light, the learned kernels can be interpreted as (noisy) unit impulses that through convolution shift the feature maps into the future according to the motion that has been observed in previous frames.  

\begin{figure}[t]
\center
\includegraphics[width=1.\linewidth]{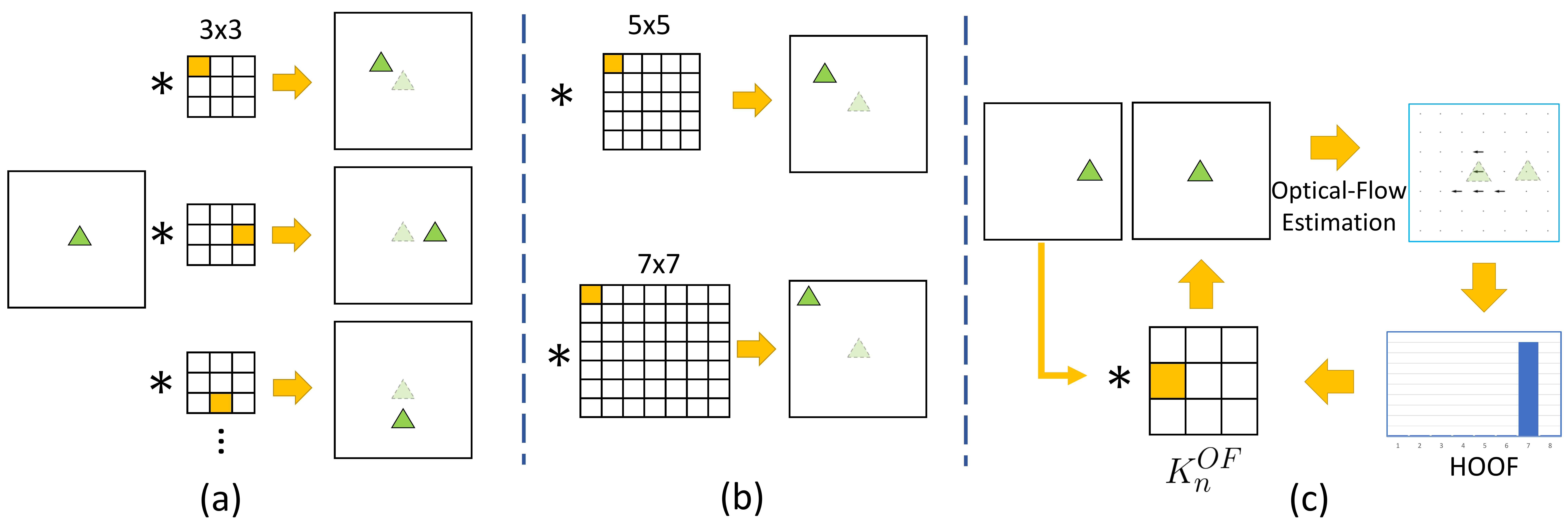}
\caption{
Illustration of spatial shifting via convolution using 2D unit impulse kernels. The positions of the impulse signal (yellow block) specifies the movement directions (a) and the displacement magnitude of the motion kernel specifies the movement stride (b). Optical flow kernels, $K^{OF}_{n}$, can be constructed to propagate frames in a convolutional manner (Section 5.1.2) and thereby can be compared to our learned motion kernels to verify that they are capturing observed motion through convolution (c).}
\label{fig:shift}
\vspace{-1em}
\end{figure}

To give more intuition, Figure~\ref{fig:shift} depicts the convolutional shifting operation with a toy example: Moving a triangle within an image using convolution. In the left subfigure, we demonstrate that there are 8 degrees of freedom in terms of motion directions for  a $3\times3$ unit impulse kernels, \ie the 8 outer entries. Once applying the convolution with any of them, the targeted triangle would be shifted accordingly by one pixel distance. Arguably, the limited shifting stride can be a shortcoming for temporal feature propagation, as it enforces a small motion assumption.
In the middle subfigure, however, it is shown that the shifting stride can be extended by expanding the size of motion kernels, \ie moving by 2 or 3 pixel distance per operation with $5\times 5$ and $7\times 7$ kernels, respectively. 
Thus, kernels with multiple scales are more flexible to capture motions that cross greater distances and also support fine-grained orientations, \eg $5\times 5$ kernels can provide 16 more directions. Our approach has adopted this multi-scale scheme.

\begin{figure}[b!]
\vspace{-2em}
\center
\includegraphics[width=1.\linewidth]{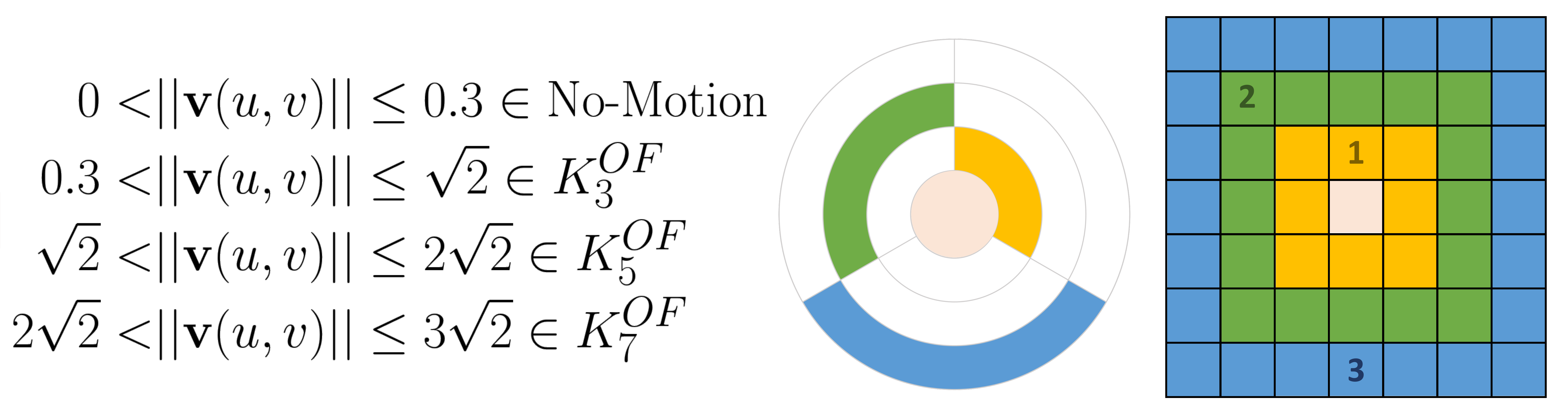}
\caption{
Discretization of flow magnitude for conversion to motion kernel format. Motion kernels are divided into 0, 1, 2 or 3 rings, for no motion, $3\times 3$, $5\times 5$ and $7\times 7$ kernels, resp., with rings corresponding to flow magnitude.
}
\label{fig:kernels}
\end{figure}

\subsubsection{Motion kernels from optical flow}
To further verify that the learned motion kernels are capturing the observed motion, we compare them to optical flow estimated from the same input frames. Apparently, no previous approach for comparing learned motion kernels with optical flow has been proposed. In response, we proceed by building on the classic Histogram of Oriented Optical Flow (HOOF) \cite{chaudhry2009histograms, lowe1999object}, with details as follows.

Let $\mathbf{v}(x,y) = (u(x,y),v(x,y))$ be optical flow, where we use a standard estimator to recover the flow \cite{farneback2003two}; although, that particular choice is not critical, \ie any reliable optical flow estimator could be used. Note that since we will be comparing to kernels that are learned on a feature channel basis, the corresponding feature channels are used as input to the flow estimator, rather than raw grey-level images. We define the flow magnitude $\| \mathbf{v}(u,v) \| = \sqrt{u^{2} + v^{2}}$ and direction $\theta = \text{tan}^{-1}(\frac{v}{u})$ in the usual way, where we suppress dependence on $(x,y)$ for compactness of notation. For conversion to the motion kernel format, we quantize the flow magnitudes and directions across $(x,y)$ into histograms with bins that correspond to the discrete kernel elements. The magnitude is quantized as shown in Figure~\ref{fig:kernels}. Note that different rings are available for quantization depending on the kernel size: $3\times 3$ kernels only encompass the inner most ring; the $5\times 5$ kernal encompasses both the inner most and middle rings; the $7\times 7$ kernel encompasses all three rings. Flow direction is discretized according to an angular range for bin $b$, out of a total of $B$ bins as
$
-\frac{\pi}{2} + \pi\frac{b-1}{B} \leq \theta < -\frac{\pi}{2} + \pi\frac{b}{B},
$
where $1 \leq b \leq B$ and $B$ is the number of orientations that can be captured by the outer ring of the motion kernel (\ie 8, 16 and 24 for motion magnitudes mapped to the inner, middle and outer rings, respectively). 
Each bin of the histogram includes designated flow vectors weighted by their magnitude, as in classic HOOF.
Finally, the histograms are smoothed with a Gaussian filter \cite{lowe1999object} and normalized to complete the conversion of optical flow to motion kernel format, yielding $K^{OF}_n$.

\subsubsection{Match results}
\begin{figure*}[t]
\includegraphics[width=1.\linewidth]{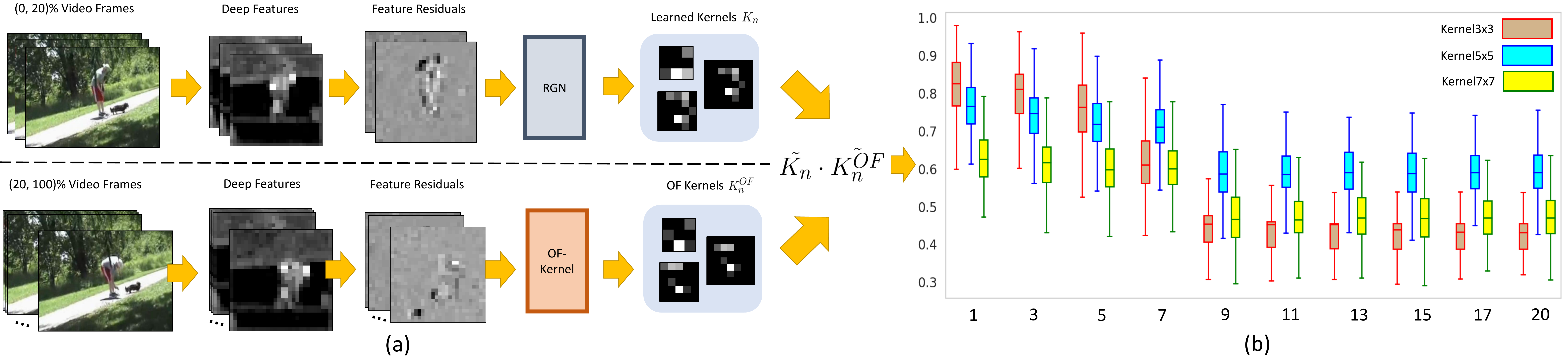}
\caption{
Pipeline for computing the match between learned motion kernels, $K_n$, and optical flow kernels, $K^{OF}_n$, for the entire prediction horizons on JHMDB21, in the deep feature residual domain, (a). Box plots of correlation results across 192 feature channels, (b). In each column, the vertical extent of the box encompasses the interquartile range, the horizontal line in the box indicates the median score and the whiskers extending above and below the box indicate the $90^{th}$ and $10^{th}$ percentiles, resp.
}
\label{fig:kernel_corr}
\end{figure*}

Figure~\ref{fig:kernel_corr}(a) illustrates our methodology for quantifying the comparison of learned motion kernels, $K_n$, with groundtruth optical flow kernels, $K^{OF}_n$, as defined in the previous section. As specified in the figure, we take the inner product between the vectorized versions of $\tilde{K_n}$ and $\tilde{K}^{OF}_n$ to quantify their match, with $\tilde{}$ denoting vectorization. Notice that since both $K_n$ and $K^{OF}_n$ are normalized by construction (Sections 3.3 and 5.1.2, resp.), their inner product is equal to their cosine similarity. In the following, we present results for all three datasets, JHMDB21, BIT and UCF101 as well as overall discussion as a series of questions and answers.


\textbf{JHMDB21.} Results for JHMDB21 are shown in Figure~\ref{fig:kernel_corr}(b) for 10 timesteps that are sampled out of all prediction horizons (\ie 22 timesteps).  It is seen that all three sets of learned kernels can demonstrate high matches for the first 7 timesteps: $3\times 3$ kernels have median correlations of $\approx$\hspace{-0.2pt}$0.8$ with OF-kernels; $5\times 5$ are at $\approx$\hspace{-0.2pt}$0.7$; $7\times 7$ are at $\approx$\hspace{-0.2pt}$0.6$. Thereafter, the matches decline  (\eg beyond the $9^{th}$ timestep, $3\times 3$ kernels can only achieve 45\%. 

It also is notable that for the $1^{st}$-$7^{th}$ prediction timesteps, the $3\times 3$ kernels achieve the highest matches and the $7\times 7$ kernels the lowest. It also is interesting to observe that the $3\times 3$ kernels degenerate more remarkably than the other two when above the $9^{th}$ prediction step. We return to both of these observations below.


\begin{figure}[t]
\centering
\includegraphics[width=1.0\linewidth]{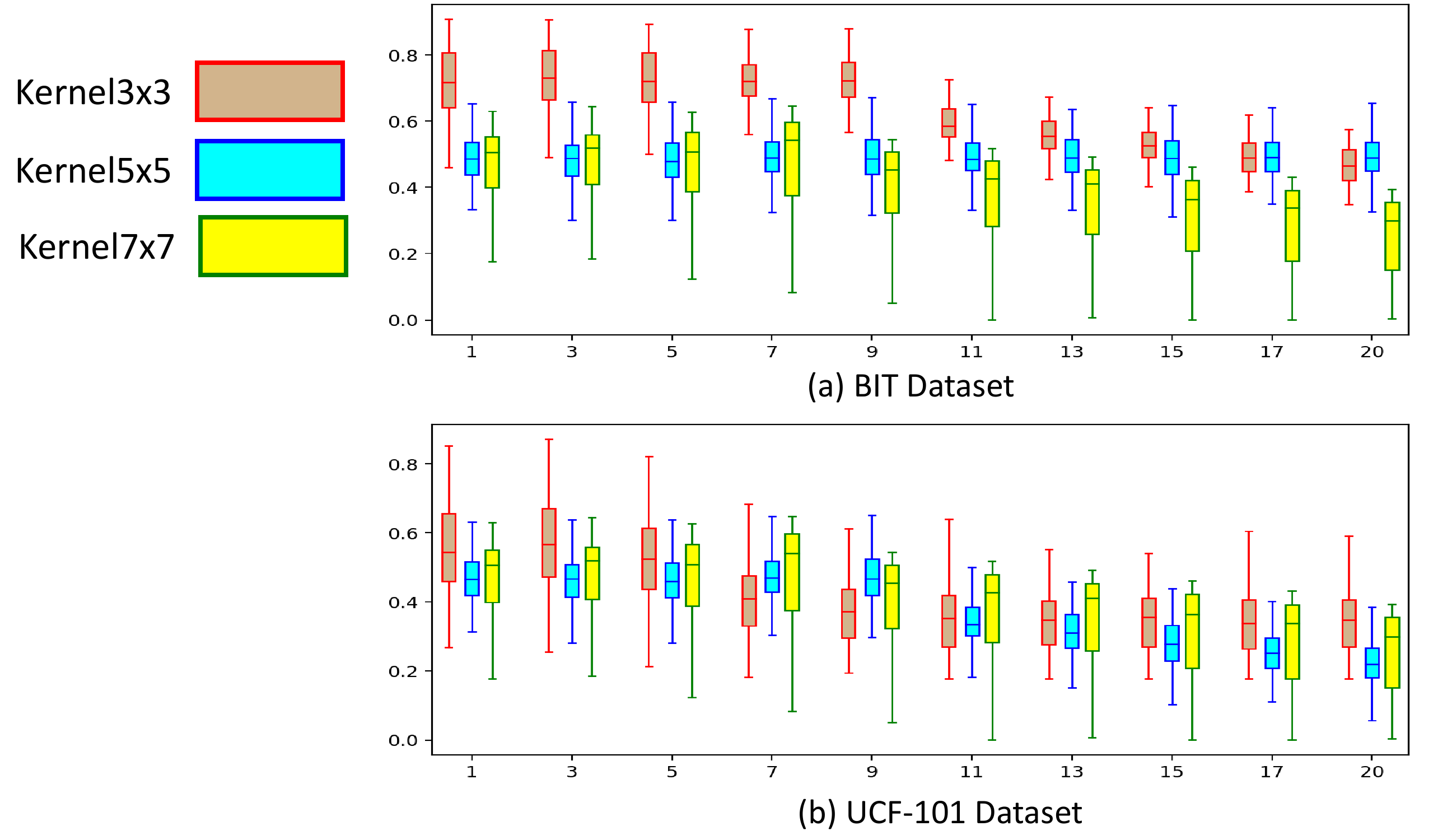}
\caption{
Box plots of correlation scores between learned kernels, $K_n$, and optical flow kernels, $K^{OF}_n$, for the BIT (a) and the UCF101 (b) datasets.
}
\label{fig:kernel_corr_extra}
\vspace{-1em}
\end{figure}

\textbf{BIT and UCF101.} Similar patterns of correlation results for BIT and UCF101 are given in Figure~\ref{fig:kernel_corr_extra}. As the BIT dataset includes mostly small actions (eg \textit{high-five}, \textit{patting} and \textit{handshaking}) and actors execute their actions smoothly, the match scores for the $3\times 3$ kernels are generally higher than the other two kernel sizes compared to that seen with JHMDB21. However, due to BIT depending on stacked optical flow input (Section~\ref{sec:empirical}), which naturally contains noise, its best correlation value is still lower than JHMDB21.
Notably, the variability of its $7\times 7$ kernels are much larger than those for JHMDB21, which likely is because the depicted slow actions do not yield large displacements that would be relevant for the largest kernel. For the UCF101 dataset, the results are not as well correlated as the others, \ie the median match scores for all three sets of kernels are lower (\eg 0.56 at the 1st timestep). These results may arise because of the greater complexity of the motions captured in UCF101, \eg arising from real-world sports video. This result also helps explain that even while our approach outperforms alternatives on this dataset (Figure~\ref{fig:result} (a)), its relative advantage compared to the other datasets is less.

\begin{figure}[t]
\centering
\includegraphics[width=1.\linewidth]{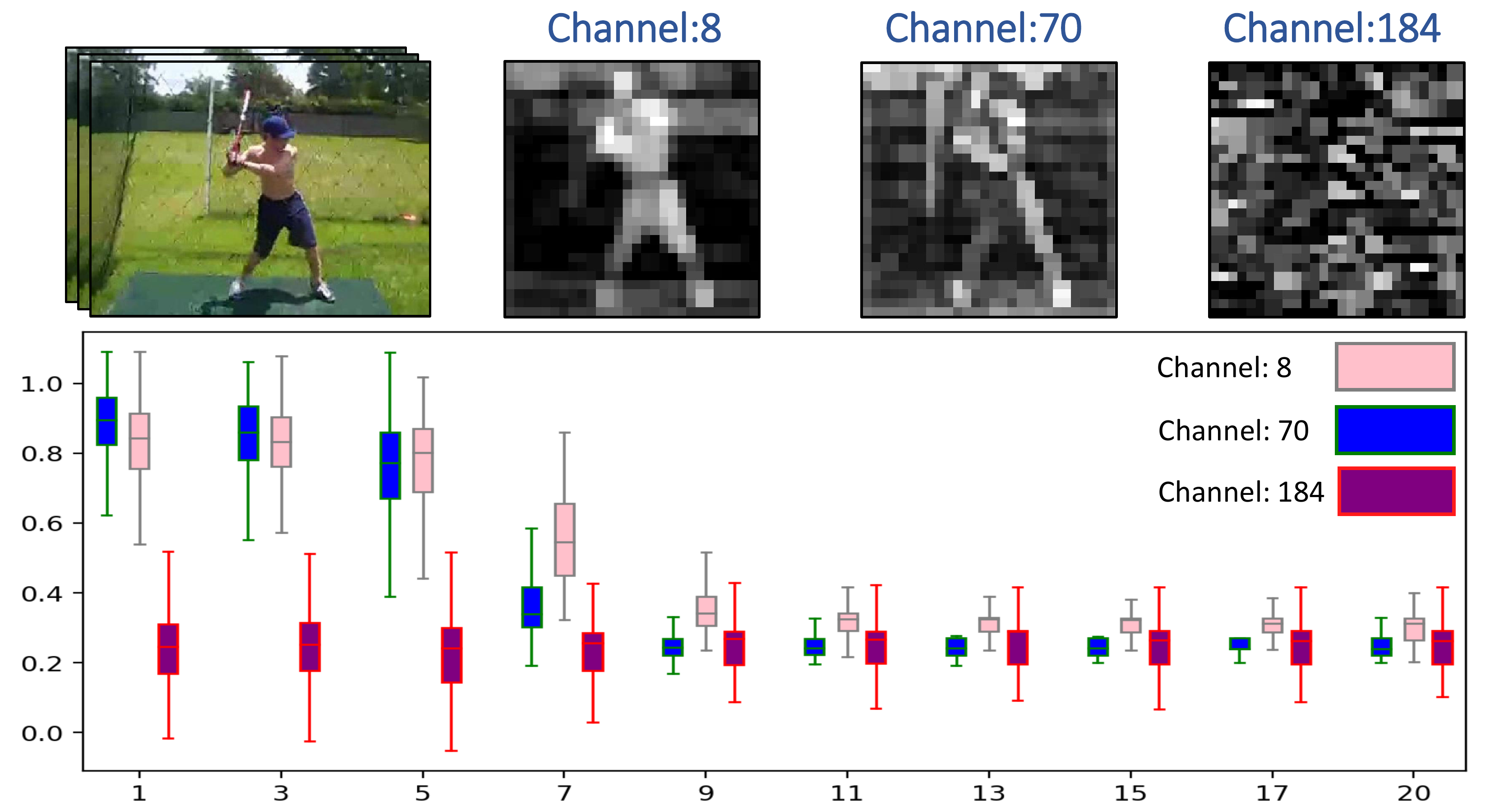}
\caption{
Visualization of three distinctive feature channels (top). Plots of the averaged matches of the learned 3x3 kernels for three selected channels across the test set (bottom). The visual quality of feature channels reflects the match scores.
}
\label{fig:fea_chn}
\vspace{-2em}
\end{figure}

\subsubsection{Discussion}\label{sec:discussion}

\textbf{Are there performance differences by feature channel?} Recall that the employed deep feature residual input to the motion kernels is a multi-channel tensor from the TSN middle stage, \ie intermediate feature with size (28, 28, 192). Our approach propagates each channel with a distinct set of kernels. Here, we visualize a few representative channels and examine the match differences between them; see Figure~\ref{fig:fea_chn}. It is seen that certain feature channels, \eg $8$ and $70$, do well in capturing the actions (\ie high feature map responses delineating the actor), while other feature maps (\eg 184), are mainly noise. Thus, their matches differ greatly: Clean channels show reasonable match scores between the learned and groundtruth kernels, whereas noisy channels show very low values throughout the predictions.
\begin{figure}[h]
\centering
\includegraphics[width=.9\linewidth]{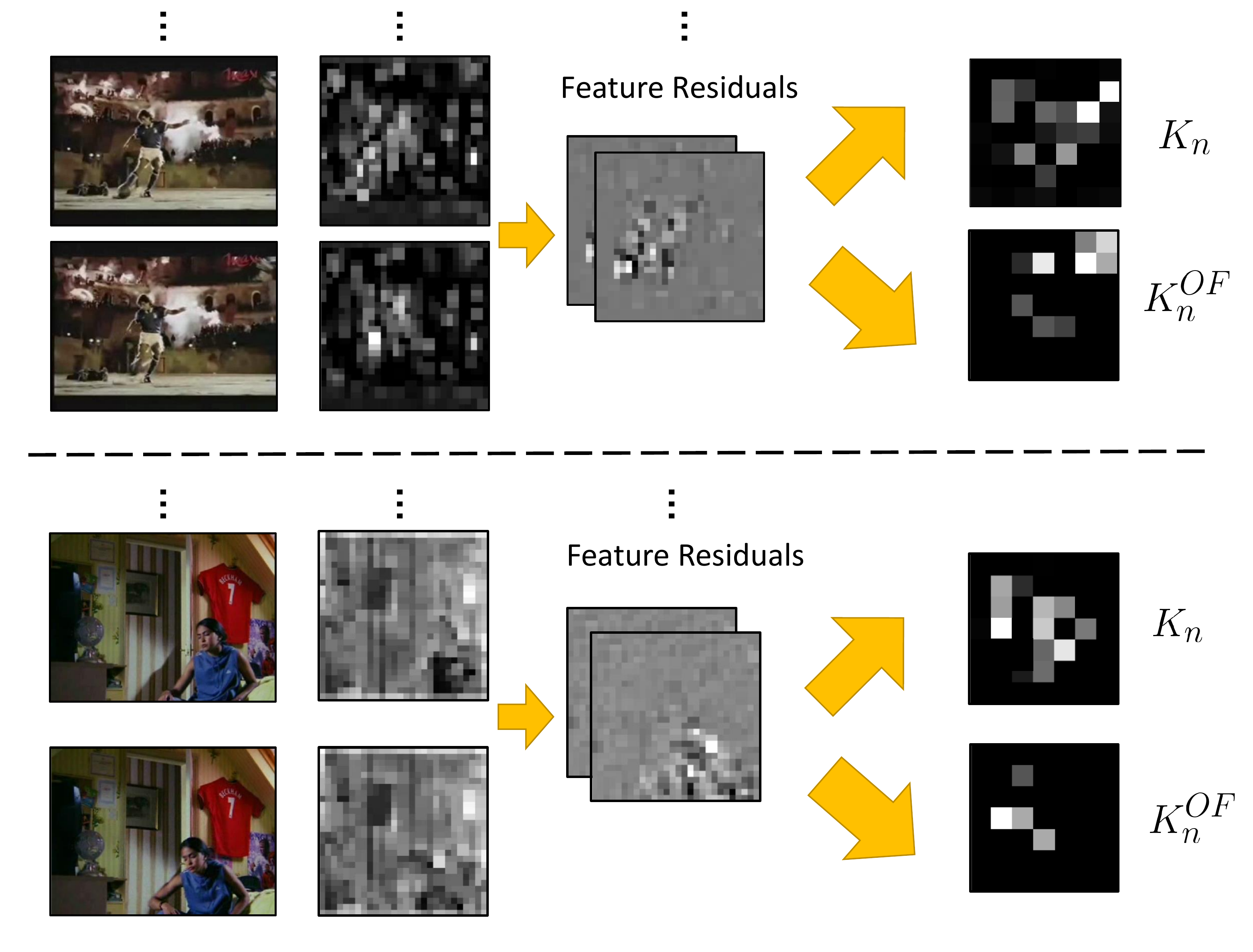}
\caption{
Plots of kernels for action videos whose deep feature residuals have flow magnitudes larger than 2 (top) and smaller than 2 (bottom).
}
\label{fig:kernel7}
\end{figure}

\textbf{Is it necessary to use larger kernel sizes?} From the $7\times 7$ kernel plots in Figure~\ref{fig:kernel_corr}(b), it is seen that these kernels have relative low correlation scores even at early timesteps, which raises doubt regarding their usefulness. Those results aggregate across all frames and actions. Examination of particular cases, however, shows the importance of larger kernels. Figure~\ref{fig:kernel7} provides an illustrative comparison. The top row depicts a \textit{kicking} action where the execution leads to a relatively large displacement (\ie half of its pixels exhibit flow magnitude greater than 2) and our learned $7\times 7$ kernels have high valued impulses at the 3 pixel displacement positions (the outermost   ring). In the bottom row, however, the \textit{standing} action executes slowly and the learned kernels reject having high values at the outermost ring. By our counting, $21$\% of deep feature residuals across the entire test set encompass flow magnitude larger than 2 and these are captured best by the $7\times7$ kernels.

\begin{figure}[h]
\centering
\includegraphics[width=1.\linewidth]{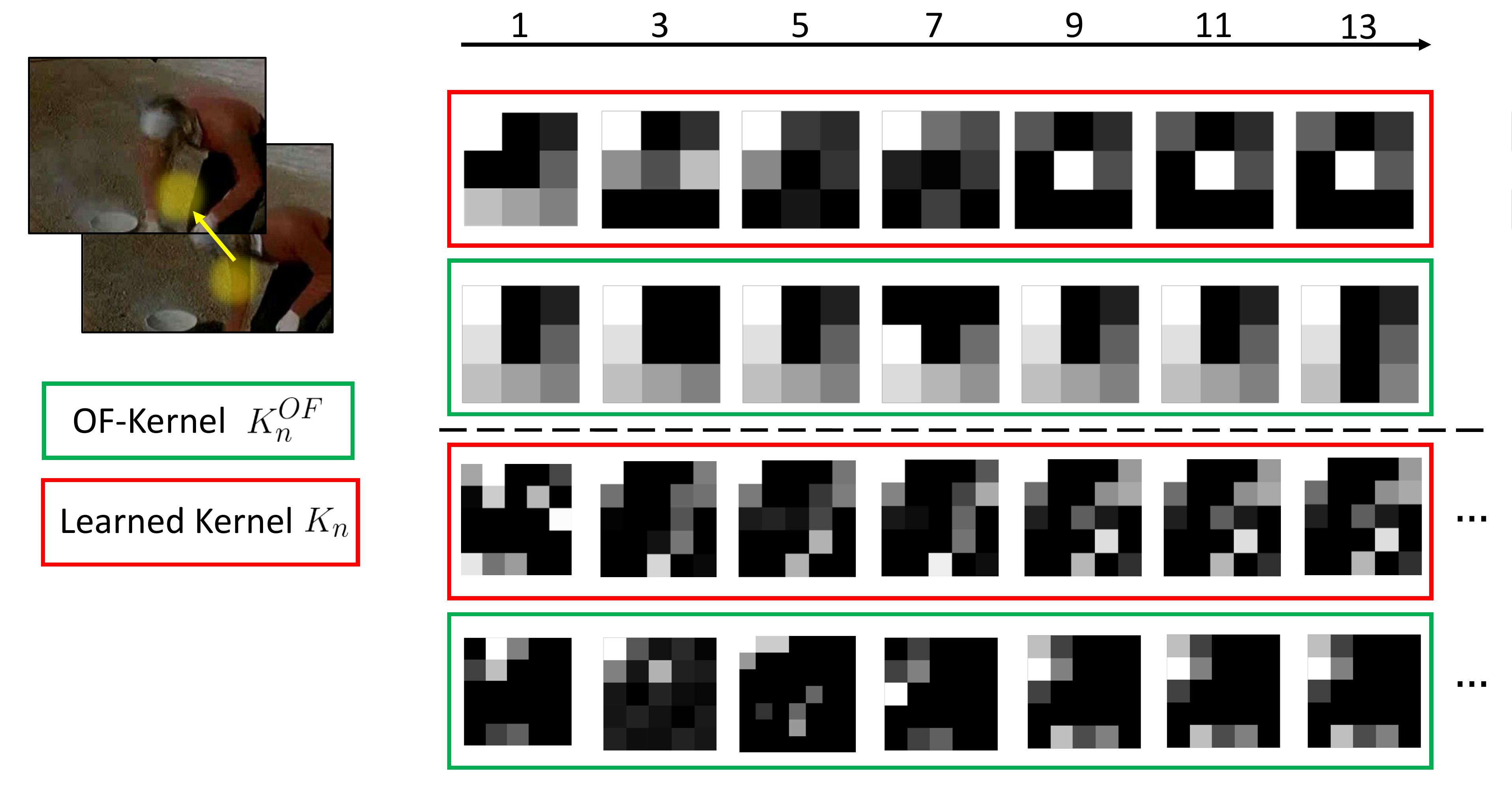}
\caption{
Visualization of failure modes of 3x3 and 5x5 kernels on a \textit{picking} test action video.
}
\label{fig:failure}
\vspace{-1em}
\end{figure}

\textbf{What are the failure modes of the learned motion kernels?} Another observation from Figure~\ref{fig:kernel_corr}(b) is that the $3\times 3$ kernels tend to degenerate more severely at longer prediction horizons than the other two, \eg its correlation goes down to $0.4$ when prediction horizon reaches the $9^{th}$ timestep and onward. To probe further, we show visualizations of a sequence of learned and groundtruth $3\times 3$ and $5\times 5$ motion kernels; see Figure~\ref{fig:failure}. It is seen that the learned $3\times 3$ kernels eventually degenerate to indicating no-motion (\ie highest response at the center position), whereas the learned $5\times 5$ kernels continue capturing the motion and thereby  yield higher correlations. This pattern is likely due to the depicted motion being relatively large and the smaller kernel has limited ability to encompass the displacement even at the beginning; this imitation is exacerbated as the shifts that it propagates lead to ever less accurate predictions at longer horizons. Thus, we see that a failure mode of the approach is capture of larger motions at longer horizons, especially as predicted by the small motion kernels. This limitation is the major one that we have discovered.

\begin{figure}[h]
\centering
\includegraphics[width=1.\linewidth]{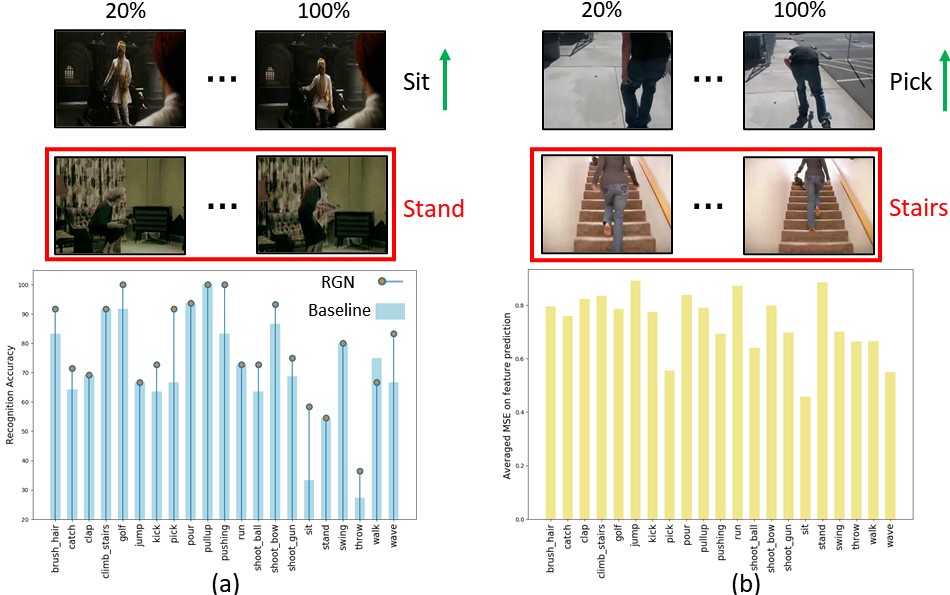}
\caption{
(Top) Visual examples for action \textit{sit} and \textit{climb stairs}, along with their most confused categories in red boxes). (Bottom) Early action recognition accuracy from our framework and baseline approach (a). Mean feature propagation error per action category (b).
}
\label{fig:confusion}
\vspace{-1em}
\end{figure}

\textbf{Is easy-to-propagate equal to easy-to-predict?} Our approach is based on the assumption that propagation into the future would bring benefits, \eg discovering the unobserved information. To examine this assumption,
we consider the relationship between accuracy improvement vs. mean feature propagation error on every action category. Results are show in Figure~\ref{fig:confusion}. The lower left subfigure shows that when being compared with the simple baseline approach (\ie using only the first 20\% of the video to predict the entire action without any further processing, as in Section 4), our approach, RGN, sees the most performance improvement on the \textit{sit} and \textit{picking} actions, \ie \textit{sit} is improved from $33.4$\% to $58.3$\% accuracy and \textit{picking} is improved from $66.7$\% to $91.7$\% accuracy. The lower right subfigure shows a plot of mean-square-error between propagated and groundtruth deep features. It is seen that the smallest error arises for \textit{sit} and \textit{pick} features, showing that these actions are relatively easy to propagate for our approach. These observations suggest that our success on action prediction is from the effectiveness of our feature propagation.


Finally, visualizations of these most improved examples and their most frequently confused categories are shown in the upper portion of Figure~\ref{fig:confusion}. It is seen that \textit{sit} can be easily confused with \textit{standing} when looking only at the static image, as they both share the same physical environments (\eg chairs). Similarly, \textit{picking} and \textit{climbing stairs} share some common subtle movements, \eg lifting-up legs. Nevertheless, our system yields greatest advantage on these examples, suggesting that its approach to predicting the future yields gains in discriminative power.


\subsection{Learned Kalman gain} 
We now examine the learned Kalman gain by plotting its values across prediction horizons. To place our results in context, we begin with a brief discussion of related work.

Injecting groundtruth mid-stage information into a recursive generation process has been seen in various previous efforts to stabilize learning, \eg scheduled sampling \cite{bengio2015scheduled} and teacher forcing \cite{williams1989learning}. Long-term sequences often suffer from error accumulation due to the poor quality of intermediate results; therefore, these approaches use real midway data at scheduled times to replace the undesired results. The Kalman filter update step serves similar purposes, but does not simply replace some predictions with real measurements: Instead of completely discarding the predictions, it corrects them proportionally to current estimates of prediction quality according to the Kalman gain. Moreover, the update is continual, rather than simply at predetermined schedule times.

\begin{figure}[h]
\centering
\includegraphics[width=1.\linewidth]{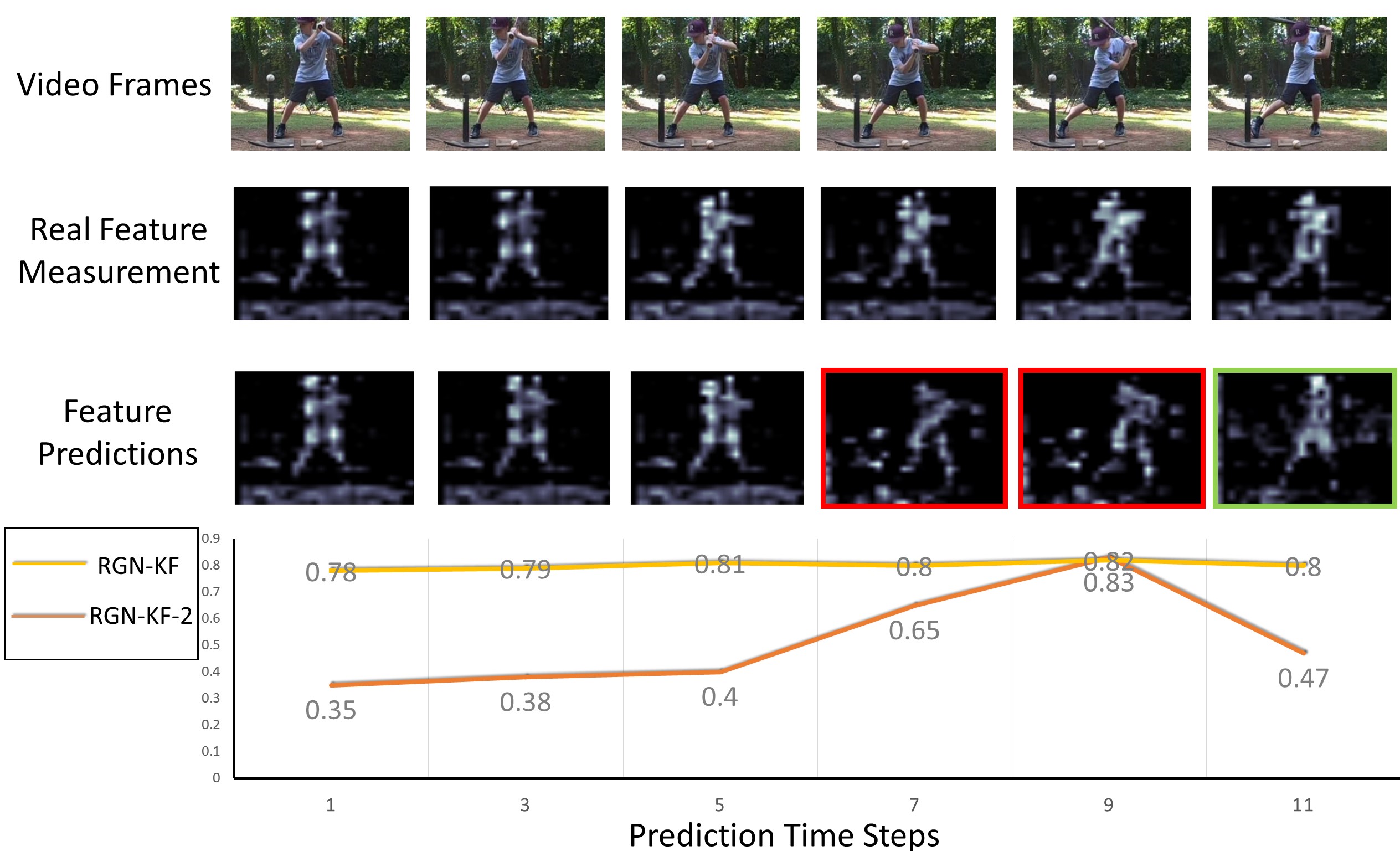}
\caption{
Illustration of learned Kalman gain values.
}
\label{fig:kalman_gain}
\end{figure}

In our approach, the Kalman gain is derived (\ie learned) from the difference between the prediction and measurement, \eqref{eq:4}, and maps to a value range $(0, 1)$, with larger values providing a larger update. By implementation, the gain update is continual and the question becomes whether it follows the prediction and measurement difference, as desired. An example is shown in Figure~\ref{fig:kalman_gain} showing an example of \textit{baseball swing}. When the prediction reaches the $7^{th}$ timestep, the propagated feature begins to degnerate (shown in red boxes) and meanwhile the Kalman gain increases, indicating that the prior estimation is less reliable. After the update at the $11^{th}$, timestep, the corrupt feature is corrected (shown in green box) and accordingly the gain value returns to a lower vale, \eg declining from $0.83$ to $0.47$. 

We further observe that for smooth actions, \eg Figure~\ref{fig:kalman_gain}, the Kalman gain starts with small values and then smoothly increases to larger values until a major update is needed and applied, after which it decreases. This behaviour follows the standard scheduled sampling strategy \cite{bengio2015scheduled} where the probability of using on-going predictions decays as timestep increases, but in our case the update is continual and is driven directly by the current state of the estimation, rather than a predefined schedule. 

\begin{figure}[t]
\centering
\includegraphics[width=1.\linewidth]{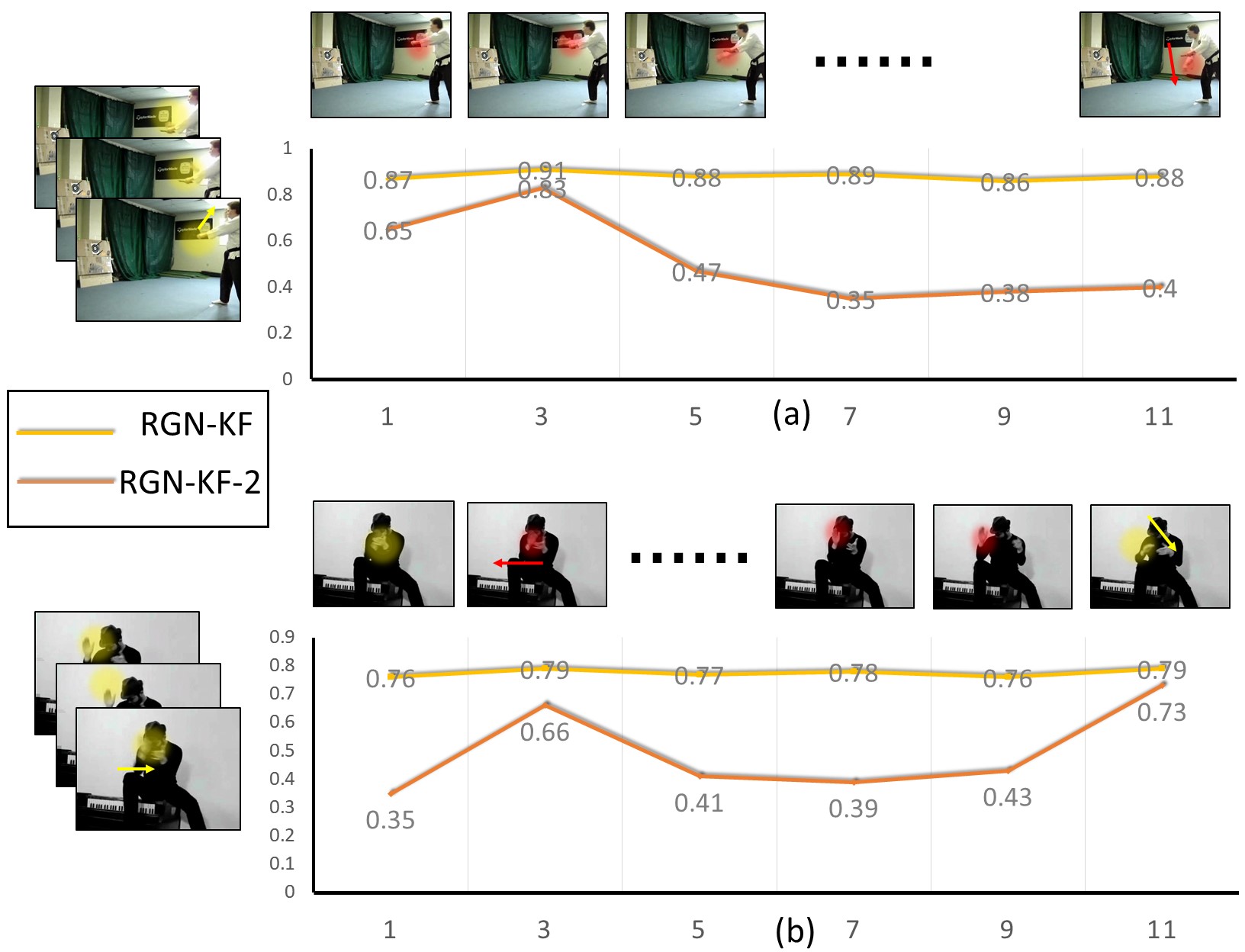}
\caption{
Visualization of learned Kalman gain of \textbf{RGN-KF-2} and \textbf{RGN-KF} on examples from the JHMDB21 dataset.
}
\label{fig:example_kalman}
\vspace{-2em}
\end{figure}

In contrast to the case of smooth motion, Figure~\ref{fig:example_kalman} illustrates performance in more complicated scenarios. Subfigure (a) depicts the situation where there is a sudden change in the direction of motion at the very beginning of the propagation and the gain immediately goes from high to low values. This behaviour relates to recent work that found reverse scheduled sampling useful for video frame prediction training \cite{wu2021predrnn}, but again our approach is adaptive to the input. Subfigure (b) provides an even more complicated example (\textit{clapping}) of multiple changes in motion directions in short time periods and the behaviour of the Kalman gain, which is seen to provide larger corrections when the direction of motion changes. These examples, illustrate that scheduling strategy should be adaptive for action videos, as there exists quite diverse motion patterns, and our Kalman filter design fullfills the need in providing updates when the motion direction changes and the prediction fails.

In addition to the plots of our current Kalman gain shown in Figures~\ref{fig:kalman_gain} and \ref{fig:example_kalman} we also show the values provided by our previous approach to learning the gain \cite{zhao2019}. As noted in Section~\ref{subsec:KF}, the earlier approach was based more simply on inputting both the prediction and observation to the learning procedure, rather than its difference. It is seen that our current approach does much better at providing corrections when most needed. In contrast, the previous approach provides uniformly high gain values, which will be prone to induce exposure bias when testing \cite{rennie2017self}.

\begin{figure}[t]
\centering
\includegraphics[width=1.\linewidth]{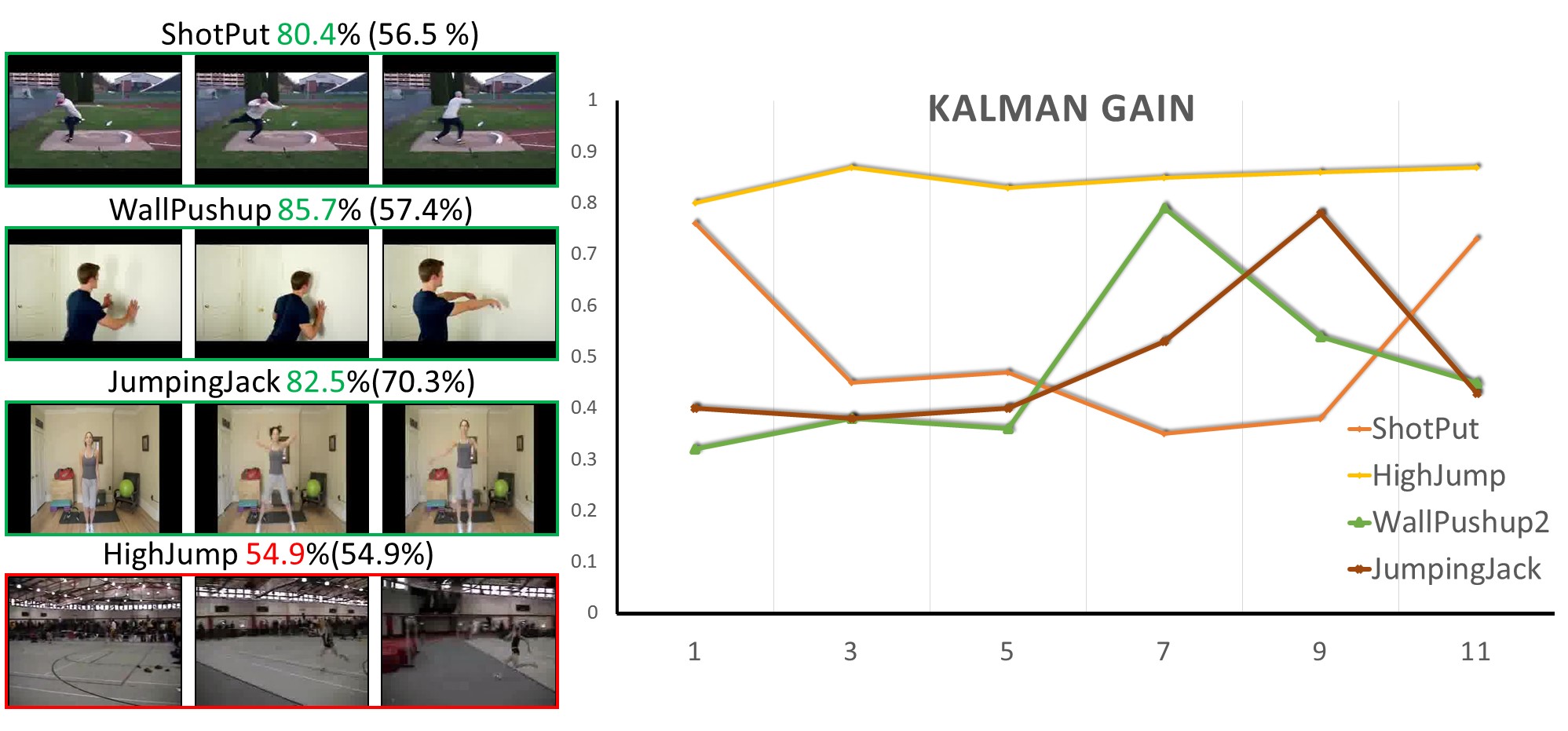}
\caption{
Visualization of changed early action recognition accuracies from \textbf{RGN-KF-2} vs. \textbf{RGN-KF} (in parentheses) on UCF101 
at observation ratio $g=0.2$ (left).  Plots of \textbf{RGN-KF-2} gain at observation ratio $g=0.2$ across time for all four classes (right).
}
\label{fig:ucf101_improve}
\vspace{-1em}
\end{figure}

Finally, we consider why the UCF101 dataset enjoys the least accuracy improvements using the new \vs previous Kalman gain formulation among the three datasets considered. We illustrate with representative examples in Figure~\ref{fig:ucf101_improve}. It seems that certain action videos, \eg \textit{ShotPut}, \textit{WallPushup} and \textit{JumpingJack}, have solid benefits and their Kalman gain values show a similar pattern as for JHMDB21 (Figures~\ref{fig:kalman_gain} and \ref{fig:example_kalman}), \ie the gain increases only at particular timesteps to correct predictions and otherwise stays relatively low. In contrast, for the \textit{HighJump} class, the gain remains uniformly high, likely due to the fact that the videos in the class show continual dramatic changes across time, \eg actors are missing at the beginning of videos and scene backgrounds drift in a substantial way due to camera motion. These changes make it necessary for \textbf{RGN-KF-2} to update with high Kalman gain across time, similar to what was seen for \textbf{RGN-KF} in Figures~\ref{fig:kalman_gain} and \ref{fig:example_kalman}; therefore, the two approaches yield similar accuracies. UCF101 tends to depict the most continual temporal changes compared to the other datasets considered, which leads the gain, and therefore final result, of the two approaches to be especially similar on this dataset. Indeed, cases where the two Kalman formulations across all datasets yield similar performance typically arise when the change within the videos is large and ongoing. Still, the newer approach is important for cases where updates are required less frequently (\eg in the top three rows of Figure~\ref{fig:ucf101_improve} as well as Figures~\ref{fig:kalman_gain} and \ref{fig:example_kalman}) and can thereby help avoid exposure bias, as discussed above.

\vspace{1cm}
\ifCLASSOPTIONcompsoc
\IEEEraisesectionheading{\section{Conclusions}\label{sec:conclusions}}
\else
\section{Conclusions}\label{sec:conclusions}
\fi
We have presented a novel spatiotemporal feature residual propagation approach to early action recognition.
Our approach learns to propagate framewise residuals in feature space to complete partial observations. The approach enjoys the advantages of the spatial structure preservation of  mid-layer ConvNet features, compact representation that captures essential information via residual processing and long-term stability via instaneous Kalman filter corrections. 
The approach has been evaluated on the UCF101, JHMDB21 and BIT-Interaction datasets, where it achieves state-of-the-art performance in comparison to a variety of alternative approaches.  We also provide a detailed investigation of what is captured by all learned components of our system to yield an overall interpretable approach.

\vspace{-1em}

\ifCLASSOPTIONcaptionsoff
  \newpage
\fi


\bibliographystyle{IEEEtran}
\bibliography{egbib.bib}{}

\begin{thebibliography}{10}
\providecommand{\url}[1]{#1}
\csname url@samestyle\endcsname
\providecommand{\newblock}{\relax}
\providecommand{\bibinfo}[2]{#2}
\providecommand{\BIBentrySTDinterwordspacing}{\spaceskip=0pt\relax}
\providecommand{\BIBentryALTinterwordstretchfactor}{4}
\providecommand{\BIBentryALTinterwordspacing}{\spaceskip=\fontdimen2\font plus
\BIBentryALTinterwordstretchfactor\fontdimen3\font minus
  \fontdimen4\font\relax}
\providecommand{\BIBforeignlanguage}[2]{{%
\expandafter\ifx\csname l@#1\endcsname\relax
\typeout{** WARNING: IEEEtran.bst: No hyphenation pattern has been}%
\typeout{** loaded for the language `#1'. Using the pattern for}%
\typeout{** the default language instead.}%
\else
\language=\csname l@#1\endcsname
\fi
#2}}
\providecommand{\BIBdecl}{\relax}
\BIBdecl

\bibitem{ryoo2011human}
M.~S. Ryoo, ``Human activity prediction: Early recognition of ongoing
  activities from streaming videos,'' in \emph{ICCV}, 2011.

\bibitem{cao2013recognize}
Y.~Cao, D.~Barrett, A.~Barbu, S.~Narayanaswamy, H.~Yu, A.~Michaux, Y.~Lin,
  S.~Dickinson, J.~Mark~Siskind, and S.~Wang, ``Recognize human activities from
  partially observed videos,'' in \emph{CVPR}, 2013.

\bibitem{lan2014hierarchical}
T.~Lan, T.-C. Chen, and S.~Savarese, ``A hierarchical representation for future
  action prediction,'' in \emph{ECCV}, 2014.

\bibitem{kong2014discriminative}
Y.~Kong, D.~Kit, and Y.~Fu, ``A discriminative model with multiple temporal
  scales for action prediction,'' in \emph{ECCV}, 2014.

\bibitem{wang2016temporal}
L.~Wang, Y.~Xiong, Z.~Wang, Y.~Qiao, D.~Lin, X.~Tang, and L.~V. Gool,
  ``Temporal segment networks: Towards good practices for deep action
  recognition,'' in \emph{ECCV}, 2016.

\bibitem{carreira2017quo}
J.~Carreira and A.~Zisserman, ``{Quo Vadis}, {Action} recognition? a new model
  and the kinetics dataset,'' in \emph{CVPR}, 2017.

\bibitem{varol2018long}
G.~Varol, I.~Laptev, and C.~Schmid, ``Long-term temporal convolutions for
  action recognition,'' \emph{IEEE Trans. PAMI}, vol.~40, no.~6, pp.
  1510--1517, 2018.

\bibitem{feichtenhofer2019slowfast}
C.~Feichtenhofer, H.~Fan, J.~Malik, and K.~He, ``Slowfast networks for video
  recognition,'' in \emph{ICCV}, 2019.

\bibitem{sadegh2017encouraging}
M.~Sadegh~Aliakbarian, F.~Sadat~Saleh, M.~Salzmann, B.~Fernando, L.~Petersson,
  and L.~Andersson, ``Encouraging {LSTMs} to anticipate actions very early,''
  in \emph{ICCV}, 2017.

\bibitem{kong2018action}
Y.~Kong, S.~Gao, B.~Sun, and Y.~Fu, ``Action prediction from videos via
  memorizing hard-to-predict samples,'' in \emph{AAAI}, 2018.

\bibitem{shi2018action}
Y.~Shi, B.~Fernando, and R.~Hartley, ``Action anticipation with {RBF}
  kernelized feature mapping {RNN},'' in \emph{ECCV}, 2018.

\bibitem{vondrick2016anticipating}
C.~Vondrick, H.~Pirsiavash, and A.~Torralba, ``Anticipating visual
  representations from unlabeled video,'' in \emph{CVPR}, 2016.

\bibitem{tung2018deep}
F.~Tung and G.~Mori, ``Deep neural network compression by in-parallel
  pruning-quantization,'' \emph{IEEE transactions on pattern analysis and
  machine intelligence}, vol.~42, no.~3, pp. 568--579, 2018.

\bibitem{zhu2017deep}
X.~Zhu, Y.~Xiong, J.~Dai, L.~Yuan, and Y.~Wei, ``Deep feature flow for video
  recognition,'' in \emph{CVPR}, 2017.

\bibitem{zeiler2014visualizing}
M.~D. Zeiler and R.~Fergus, ``Visualizing and understanding convolutional
  networks,'' in \emph{ECCV}, 2014.

\bibitem{zhao2019}
H.~Zhao and R.~P. Wildes, ``Spatiotemporal feature residual propagation for
  action prediction,'' in \emph{ICCV}, 2019.

\bibitem{ryoo2009spatio}
M.~S. Ryoo and J.~K. Aggarwal, ``Spatio-temporal elationship match: Video
  structure comparison for recognition of complex human activities,'' in
  \emph{ICCV}, 2009.

\bibitem{singh2017online}
G.~Singh, S.~Saha, M.~Sapienza, P.~H. Torr, and F.~Cuzzolin, ``Online real-time
  multiple spatiotemporal action localisation and prediction,'' in \emph{ICCV},
  2017.

\bibitem{singh2018predicting}
G.~Singh, S.~Saha, and F.~Cuzzolin, ``Predicting action tubes,'' in
  \emph{ECCVW}, 2018.

\bibitem{kong2017deep}
Y.~Kong, Z.~Tao, and Y.~Fu, ``Deep sequential context networks for action
  prediction,'' in \emph{CVPR}, 2017.

\bibitem{kong2018adversarial}
------, ``Adversarial action prediction networks,'' \emph{IEEE transactions on
  pattern analysis and machine intelligence}, vol.~42, no.~3, pp. 539--553,
  2018.

\bibitem{gammulle2019predicting}
H.~Gammulle, S.~Denman, S.~Sridharan, and C.~Fookes, ``Predicting the future: A
  jointly learnt model for action anticipation,'' in \emph{ICCV}, 2019.

\bibitem{wang2019progressive}
X.~Wang, J.-F. Hu, J.-H. Lai, J.~Zhang, and W.-S. Zheng, ``Progressive
  teacher-student learning for early action prediction,'' in \emph{CVPR}, 2019.

\bibitem{li2020hard}
T.~Li, J.~Liu, W.~Zhang, and L.~Duan, ``Hard-net: {Hardness-aware}
  discrimination network for {3D} early activity prediction,'' in \emph{ECCV},
  2020.

\bibitem{wu2021spatial}
X.~Wu, R.~Wang, J.~Hou, H.~Lin, and J.~Luo, ``Spatial--temporal relation
  reasoning for action prediction in videos,'' \emph{IJCV}, pp. 1--22, 2021.

\bibitem{chen2020group}
J.~Chen, W.~Bao, and Y.~Kong, ``Group activity prediction with sequential
  relational anticipation model,'' in \emph{ECCV}, 2020.

\bibitem{xue2016visual}
T.~Xue, J.~Wu, K.~L. Bouman, and W.~T. Freeman, ``Visual dynamics:
  Probabilistic future frame synthesis via cross convolutional networks,''
  \emph{arXiv preprint arXiv:1607.02586}, 2016.

\bibitem{jia2016dynamic}
X.~Jia, B.~De~Brabandere, T.~Tuytelaars, and L.~Van~Gool, ``Dynamic filter
  networks,'' in \emph{NIPS}, 2016.

\bibitem{dai2017deformable}
J.~Dai, H.~Qi, Y.~Xiong, Y.~Li, G.~Zhang, H.~Hu, and Y.~Wei, ``Deformable
  convolutional networks,'' in \emph{ICCV}, 2017.

\bibitem{bako2017kernel}
S.~Bako, T.~Vogels, B.~McWilliams, M.~Meyer, J.~Nov{\'a}k, A.~Harvill, P.~Sen,
  T.~Derose, and F.~Rousselle, ``Kernel-predicting convolutional networks for
  denoising monte carlo renderings.'' \emph{ACM Trans. Graph.}, vol.~36, no.~4,
  pp. 97--1, 2017.

\bibitem{pouyanfar2018dynamic}
S.~Pouyanfar, Y.~Tao, A.~Mohan, H.~Tian, A.~S. Kaseb, K.~Gauen, R.~Dailey,
  S.~Aghajanzadeh, Y.-H. Lu, S.-C. Chen \emph{et~al.}, ``Dynamic sampling in
  convolutional neural networks for imbalanced data classification,'' in
  \emph{MIPR}, 2018.

\bibitem{lan2017deep}
Z.~Lan, Y.~Zhu, A.~G. Hauptmann, and S.~Newsam, ``Deep local video feature for
  action recognition,'' in \emph{CVPRW}, 2017.

\bibitem{diba2017deep}
A.~Diba, V.~Sharma, and L.~Van~Gool, ``Deep temporal linear encoding
  networks,'' in \emph{CVPR}, 2017.

\bibitem{wang2020video}
H.~Wang, D.~Tran, L.~Torresani, and M.~Feiszli, ``Video modeling with
  correlation networks,'' in \emph{CVPR}, 2020.

\bibitem{dosovitskiy2016inverting}
A.~Dosovitskiy and T.~Brox, ``Inverting visual representations with
  convolutional networks,'' in \emph{CVPR}, 2016.

\bibitem{feichtenhofer2017spatiotemporal}
C.~Feichtenhofer, A.~Pinz, and R.~P. Wildes, ``Spatiotemporal multiplier
  networks for video action recognition,'' in \emph{CVPR}, 2017.

\bibitem{le1991mpeg}
D.~L. Gall, ``{MPEG}: A video compression standard for multimedia
  applications,'' \emph{Commun. ACM}, vol.~34, pp. 46--58, 1991.

\bibitem{Anandan89}
P.~Anandan, ``A computational framework and an algorithm for the measurement of
  visual motion,'' \emph{IJCV}, vol.~2, no.~3, pp. 283--310, 1989.

\bibitem{ranjan2017optical}
A.~Ranjan and M.~J. Black, ``Optical flow estimation using a spatial pyramid
  network,'' in \emph{CVPR}, 2017.

\bibitem{jiao2017formresnet}
J.~Jiao, W.-C. Tu, S.~He, and R.~W.~H. Lau, ``{FormResNet}: Formatted residual
  learning for image restoration,'' in \emph{CVPRW}, 2017.

\bibitem{lu2018deep}
G.~Lu, W.~Ouyang, D.~Xu, X.~Zhang, Z.~Gao, and M.-T. Sun, ``Deep {Kalman}
  filtering network for video compression artifact reduction,'' in \emph{ECCV},
  2018.

\bibitem{wu2018compressed}
C.-Y. Wu, M.~Zaheer, H.~Hu, R.~Manmatha, A.~J. Smola, and
  P.~Kr{\"a}henb{\"u}hl, ``Compressed video action recognition,'' in
  \emph{CVPR}, 2018.

\bibitem{kalman1960new}
R.~E. Kalman, ``A new approach to linear filtering and prediction problems,''
  \emph{Journal of Basic Engineering}, vol.~82, no.~1, pp. 35--45, 1960.

\bibitem{dave2017predictive}
A.~Dave, O.~Russakovsky, and D.~Ramanan, ``Predictive-corrective networks for
  action detection,'' in \emph{CVPR}, 2017.

\bibitem{guen2020disentangling}
V.~L. Guen and N.~Thome, ``Disentangling physical dynamics from unknown factors
  for unsupervised video prediction,'' in \emph{CVPR}, 2020.

\bibitem{haarnoja2016backprop}
T.~Haarnoja, A.~Ajay, S.~Levine, and P.~Abbeel, ``Backprop {KF}: Learning
  discriminative deterministic state estimators,'' \emph{arXiv preprint
  arXiv:1605.07148}, 2016.

\bibitem{coskun2017long}
H.~Coskun, F.~Achilles, R.~DiPietro, N.~Navab, and F.~Tombari, ``Long
  short-term memory {Kalman} filters: Recurrent neural estimators for pose
  regularization,'' in \emph{ICCV}, 2017.

\bibitem{tran2015learning}
D.~Tran, L.~Bourdev, R.~Fergus, L.~Torresani, and M.~Paluri, ``Learning
  spatiotemporal features with {3D} convolutional networks,'' in \emph{ICCV},
  2015.

\bibitem{zhou2016learning}
B.~Zhou, A.~Khosla, A.~Lapedriza, A.~Oliva, and A.~Torralba, ``Learning deep
  features for discriminative localization,'' in \emph{CVPR}, 2016.

\bibitem{feichtenhofer2020deep}
C.~Feichtenhofer, A.~Pinz, R.~P. Wildes, and A.~Zisserman, ``Deep insights into
  convolutional networks for video recognition,'' \emph{IJCV}, vol. 128, no.~2,
  pp. 420--437, 2020.

\bibitem{gabor1946}
D.~Gabor, ``Theory of communication,'' \emph{Journal of the Institute of
  Electrical Engineers}, vol.~93, pp. 429--457, 1946.

\bibitem{luan2018gabor}
S.~Luan, C.~Chen, B.~Zhang, J.~Han, and J.~Liu, ``Gabor convolutional
  networks,'' \emph{IEEE TIP}, vol.~27, no.~9, pp. 4357--4366, 2018.

\bibitem{hadji2020convolutional}
I.~Hadji and R.~P. Wildes, ``Why convolutional networks learn oriented bandpass
  filters: Theory and empirical support,'' \emph{arXiv preprint
  arXiv:2011.14665}, 2020.

\bibitem{bau2020understanding}
D.~Bau, J.-Y. Zhu, H.~Strobelt, A.~Lapedriza, B.~Zhou, and A.~Torralba,
  ``Understanding the role of individual units in a deep neural network,''
  \emph{Proceedings of the National Academy of Sciences}, vol. 117, no.~48, pp.
  30\,071--30\,078, 2020.

\bibitem{monga2021algorithm}
V.~Monga, Y.~Li, and Y.~C. Eldar, ``Algorithm unrolling: Interpretable,
  efficient deep learning for signal and image processing,'' \emph{IEEE Signal
  Processing Magazine}, vol.~38, no.~2, pp. 18--44, 2021.

\bibitem{finn2016unsupervised}
C.~Finn, I.~Goodfellow, and S.~Levine, ``Unsupervised learning for physical
  interaction through video prediction,'' in \emph{NIPS}, 2016.

\bibitem{reda2018sdc}
F.~A. Reda, G.~Liu, K.~J. Shih, R.~Kirby, J.~Barker, D.~Tarjan, A.~Tao, and
  B.~Catanzaro, ``{SDCNet}: Video prediction using spatially-displaced
  convolution,'' in \emph{ECCV}, 2018.

\bibitem{mathieu2015deep}
M.~Mathieu, C.~Couprie, and Y.~LeCun, ``Deep multi-scale video prediction
  beyond mean square error,'' \emph{arXiv preprint arXiv:1511.05440}, 2015.

\bibitem{villegas2017decomposing}
R.~Villegas, J.~Yang, S.~Hong, X.~Lin, and H.~Lee, ``Decomposing motion and
  content for natural video sequence prediction,'' in \emph{ICLR}, 2017.

\bibitem{byeon2018contextvp}
W.~Byeon, Q.~Wang, R.~K. Srivastava, and P.~Koumoutsakos, ``Contextvp: Fully
  context-aware video prediction,'' in \emph{ECCV}, 2018.

\bibitem{chen2012marginalized}
M.~Chen, Z.~Xu, K.~Weinberger, and F.~Sha, ``Marginalized denoising
  autoencoders for domain adaptation,'' \emph{arXiv preprint arXiv:1206.4683},
  2012.

\bibitem{soomro2012ucf101}
K.~Soomro, A.~R. Zamir, and M.~Shah, ``{UCF101}: A dataset of 101 human actions
  classes from videos in the wild,'' \emph{arXiv preprint arXiv:1212.0402},
  2012.

\bibitem{jhuang2013towards}
H.~Jhuang, J.~Gall, S.~Zuffi, C.~Schmid, and M.~J. Black, ``Towards
  understanding action recognition,'' in \emph{ICCV}, 2013.

\bibitem{kong2014interactive}
Y.~Kong, Y.~Jia, and Y.~Fu, ``Interactive phrases: Semantic descriptions for
  human interaction recognition,'' \emph{IEEE Trans. PAMI}, vol.~36, no.~9, pp.
  1775--1788, 2014.

\bibitem{kuehne2011hmdb}
H.~Kuehne, H.~Jhuang, E.~Garrote, T.~Poggio, and T.~Serre, ``{HMDB}: A large
  video database for human motion recognition,'' in \emph{ICCV}, 2011.

\bibitem{he2016deep}
K.~He, X.~Zhang, S.~Ren, and J.~Sun, ``Deep residual learning for image
  recognition,'' in \emph{CVPR}, 2016.

\bibitem{glorot2011deep}
X.~Glorot, A.~Bordes, and Y.~Bengio, ``Deep sparse rectifier neural networks,''
  in \emph{AISTATS}, 2011.

\bibitem{paszke2017automatic}
A.~Paszke, S.~Gross, S.~Chintala, G.~Chanan, E.~Yang, Z.~DeVito, Z.~Lin,
  A.~Desmaison, L.~Antiga, and A.~Lerer, ``Automatic differentiation in
  {PyTorch},'' in \emph{NIPS Autodiff Workshop}, 2017.

\bibitem{ma2016learning}
S.~Ma, L.~Sigal, and S.~Sclaroff, ``Learning activity progression in lstms for
  activity detection and early detection,'' in \emph{CVPR}, 2016.

\bibitem{soomro2019online}
K.~Soomro, H.~Idrees, and M.~Shah, ``Online localization and prediction of
  actions and interactions,'' \emph{IEEE Trans. PAMI}, vol.~41, no.~2, pp.
  459--472, 2018.

\bibitem{soomro2016predicting}
------, ``Predicting the where and what of actors and actions through online
  action localization,'' in \emph{CVPR}, 2016.

\bibitem{jain2016recurrent}
A.~Jain, A.~Singh, H.~S. Koppula, S.~Soh, and A.~Saxena, ``Recurrent neural
  networks for driver activity anticipation via sensory-fusion architecture,''
  in \emph{ICRA}, 2016.

\bibitem{xingjian2015convolutional}
X.~Shi, Z.~Chen, H.~Wang, D.-Y. Yeung, W.-K. Wong, and W.-C. Woo,
  ``Convolutional {LSTM} network: A machine learning approach for precipitation
  nowcasting,'' in \emph{NIPS}, 2015.

\bibitem{bracewell1986}
R.~N. Bracewell, \emph{The Fourier Transform and its Applications}.\hskip 1em
  plus 0.5em minus 0.4em\relax McGraw Hill, 1986.

\bibitem{chaudhry2009histograms}
R.~Chaudhry, A.~Ravichandran, G.~Hager, and R.~Vidal, ``Histograms of oriented
  optical flow and binet-cauchy kernels on nonlinear dynamical systems for the
  recognition of human actions,'' in \emph{CVPR}, 2009.

\bibitem{lowe1999object}
D.~G. Lowe, ``Object recognition from local scale-invariant features,'' in
  \emph{ICCV}, 1999.

\bibitem{farneback2003two}
G.~Farneb{\"a}ck, ``Two-frame motion estimation based on polynomial
  expansion,'' in \emph{SCIA}, 2003.

\bibitem{bengio2015scheduled}
S.~Bengio, O.~Vinyals, N.~Jaitly, and N.~Shazeer, ``Scheduled sampling for
  sequence prediction with recurrent neural networks,'' in \emph{NIPS}, 2015.

\bibitem{williams1989learning}
R.~J. Williams and D.~Zipser, ``A learning algorithm for continually running
  fully recurrent neural networks,'' \emph{Neural computation}, vol.~1, no.~2,
  pp. 270--280, 1989.

\bibitem{wu2021predrnn}
H.~Wu, J.~Zhang, Z.~Gao, J.~Wang, P.~S. Yu, M.~Long \emph{et~al.}, ``{PredRNN}:
  A recurrent neural network for spatiotemporal predictive learning,''
  \emph{arXiv preprint arXiv:2103.09504}, 2021.

\bibitem{rennie2017self}
S.~J. Rennie, E.~Marcheret, Y.~Mroueh, J.~Ross, and V.~Goel, ``Self-critical
  sequence training for image captioning,'' in \emph{CVPR}, 2017.

\end{thebibliography}

\vspace{-1.25cm}
\begin{IEEEbiography}[{\includegraphics[width=1in,height=1.25in,clip,keepaspectratio]{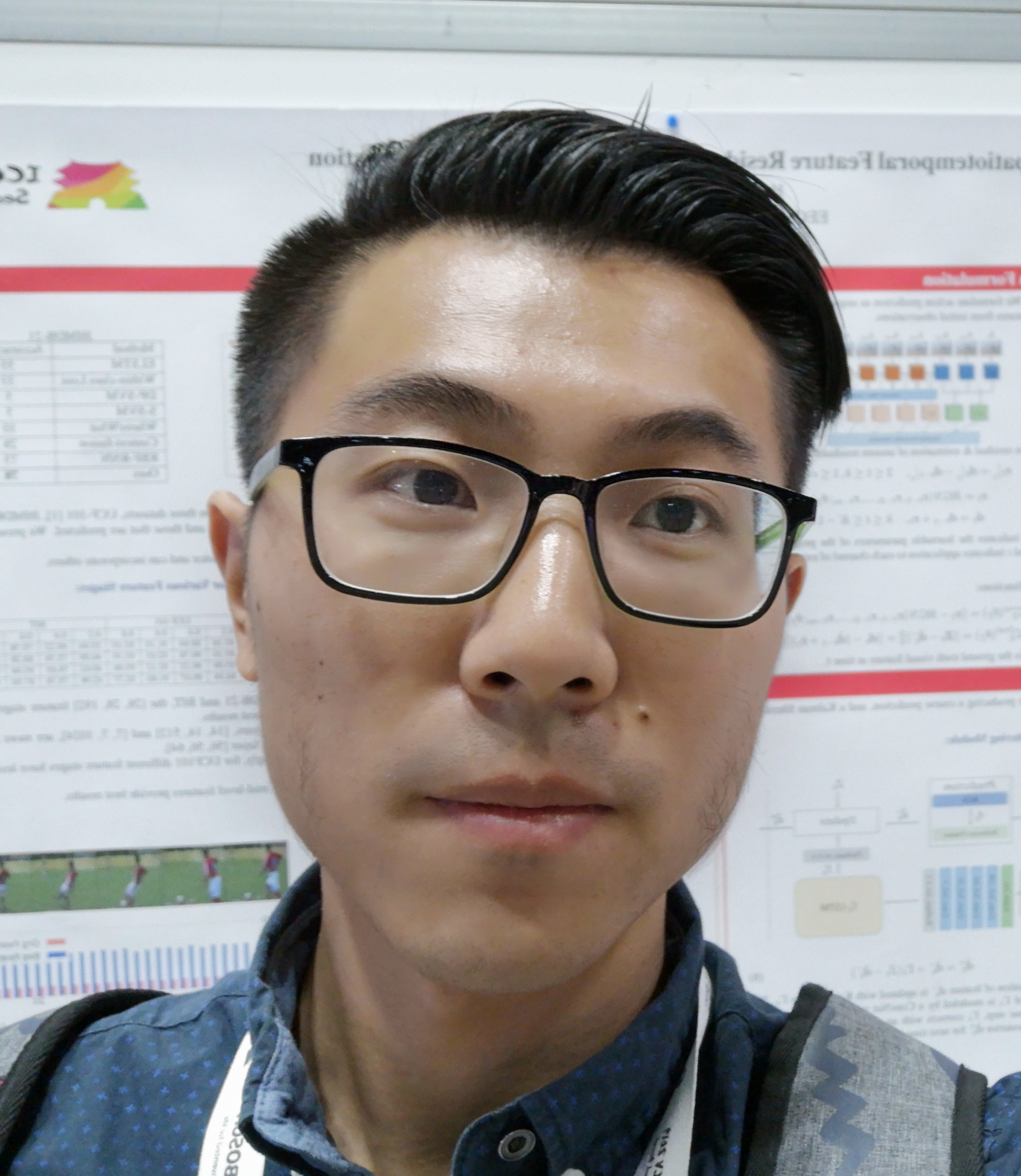}}]{He Zhao}
received the B. Eng. degree in communication engineering from ZhengZhou University,
China, in 2014, and the MSc degree in computer science from the University of Florida, Gainesville, Florida, USA, in 2015. Currently, he is pursuing the doctoral degree in the Department of Electrical Engineering and Computer Science at York University, Toronto, Canada. He is a recipient an Ontario Trillium Foundation graduate scholarship. His
major field of interest is computer vision with an emphasis on video dynamics and human motion understanding.
\end{IEEEbiography}
\vspace{-1cm}
\begin{IEEEbiography}    [{\includegraphics[width=1in,height=1.25in,clip,keepaspectratio]{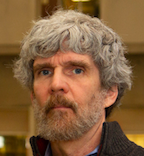}}]
{Richard P. Wildes}
 (Member, IEEE) received
the PhD degree from the Massachusetts Institute of Technology in 1989. Subsequently,
he joined Sarnoff Corporation in Princeton,
New Jersey, as a Member of the Technical Staff in the Vision Technologies Lab. In
2001, he joined the Department of Electrical
Engineering and Computer Science at York
University, Toronto, where he is a
Professor, a member of the Centre for
Vision Research and a Tier I York Research Chair. Honours include receiving
a Sarnoff Corporation Technical Achievement Award, the IEEE D.G.
Fink Prize Paper Award and twice giving
invited presentations to the US National Academy of Sciences.
His main areas of research interest are computational vision, especially video understanding, and artificial
intelligence.
\end{IEEEbiography}


\end{document}